\documentclass[journal]{IEEEtran}

\usepackage{cite}

\usepackage{graphicx}
\ifCLASSINFOpdf

\else

\fi

\usepackage{booktabs}  
\usepackage{multirow}  
\usepackage{amssymb}

\usepackage{amsmath}

\usepackage[final,hidelinks]{hyperref}

\usepackage{url}

\begin{document}

\title{WoVR+: World Model as Reliable Simulator for Versatile VLA Policy Optimization with RL}

\author{
\mbox{Shangqing Zhou\textsuperscript{1,2,\ensuremath{\dagger}}},
\mbox{Zhennan Jiang\textsuperscript{1,2,3,\ensuremath{\dagger}}},
\mbox{Ning Ma\textsuperscript{4}},
\mbox{Xin Fu\textsuperscript{2}},
\mbox{Yupeng Zheng\textsuperscript{1,2}},
\mbox{Shuai Tian\textsuperscript{1,2}},
\mbox{Huangrui Li\textsuperscript{1,2}},
\mbox{Jiahao Liu\textsuperscript{1}},
\mbox{Chao Yu\textsuperscript{5}},
\mbox{Zhongpu Xia\textsuperscript{1,6}},
\mbox{Haoran Li\textsuperscript{1,*}},
\mbox{Dongbin Zhao\textsuperscript{1}}
\thanks{
\textsuperscript{1} Institute of Automation, Chinese Academy of Sciences.
}
\thanks{
\textsuperscript{2} School of Artificial Intelligence, University of Chinese Academy of Sciences.
}
\thanks{
\textsuperscript{3} Zhongguancun Academy.
}
\thanks{
\textsuperscript{4} Xi'an Jiaotong-Liverpool University.
}
\thanks{
\textsuperscript{5} Tsinghua University.
}
\thanks{
\textsuperscript{6} Anyverse Dynamics.
}
\thanks{
\ensuremath{\dagger} Shangqing Zhou and Zhennan Jiang contributed equally to this work.
}
\thanks{
* Haoran Li is the corresponding author.
}
}

\markboth{}%
{}

\maketitle


\begin{abstract}
Action-conditioned world models can reduce costly real-robot interaction in reinforcement learning for Vision-Language-Action models (VLA-RL), but hallucinations can mislead policy optimization.
Extending them to richer VLA interfaces and long-horizon tasks raises challenges in spatiotemporal and cross-modal consistency, exploration–reliability balance in RL, and compounding rollout errors.
We present WoVR+, extending WoVR's hallucination-aware WM-RL framework by redesigning the learned environment and imagined interaction protocol:
A VLA-interface-aligned world model provides multi-view prediction and proprioceptive updates while mitigating hallucinations.
Chunk-wise noise preserves flow-based policy exploration while limiting prediction-degrading action jitter.
For long-horizon tasks, stage-local imagined training combines stage-entry resets, stage-specific rewards, local horizons, and controllable sampling to shorten generated prefixes and balance training across stages.
WoVR+ improves success over SFT by 21.09\% and 16.7\% on RoboTwin and 6 real-robot tasks. Long-horizon phone-packaging success rises from 30\% to 75\%.
Videos are available at \href{https://youmo445.github.io/WoVR_Plus.github.io/}{\nolinkurl{https://youmo445.github.io/WoVR_Plus.github.io/}}.
\end{abstract}


\IEEEpeerreviewmaketitle

\section{Introduction}

\IEEEPARstart{V}{ision}-Language-Action (VLA) models~\cite{openvlaoft,pi05} are playing an increasingly important role in robotic manipulation~\cite{robotic_foundation_model}, where actions are generated end-to-end by conditioning on language instructions and visual observations. Most existing VLA systems are trained via supervised fine-tuning (SFT). Although effective in many downstream tasks, this paradigm fundamentally limits the performance ceiling of VLA policies, as it is tightly constrained by the quality and coverage of demonstration data.

Reinforcement Learning (RL)~\cite{rl100} can further improve VLA  policies~\cite{rlinfvla,rlbringvla} obtained through SFT. However, off-policy methods are generally more sample-efficient but often suffer from distribution shift and training instability~\cite{conrft}, whereas on-policy updates~\cite{ppo, deepseekmath} require massive parallel environment interaction for stable and efficient training~\cite{chen2026pitextttrlonlinerlfinetuning}, making them impractical for real robots due to costly data collection~\cite{hilserl}. Although simulation-based alternatives have been explored~\cite{simplevlarl, DexSim2Real2,Simulation-Aided}, aligning simulators with real-world dynamics remains highly challenging.

Meanwhile, Action-conditioned world models (ACWMs)~\cite{hunyuan2,yume1_5,lingbotwm} based on video generation models~\cite{wan2025wanopenadvancedlargescale,opensora} offer the possibility of supporting large-scale imagined interaction with limited robot data.
Several works directly treat them as learned environments~\cite{WMPO,xiao2026worldenvleveragingworldmodel,jiang2026world4rldiffusionworldmodels} to optimize VLA policies through RL~\cite{liu2026worldvlaloopclosedlooplearningvideo,zhang2026reinforcingactionpoliciesprophesying}, using imagined interactions to reduce additional online robot interaction.

However, learned ACWMs are not faithful simulators~\cite{jiang2026wovrworldmodelsreliable,zhang2026practicalworldmodelbasedreinforcement}. Here, \emph{hallucination} refers to a systematic mismatch between imagined and real outcomes in closed-loop interaction. ACWMs may generate visually plausible yet physically incorrect rollouts, resulting in spurious success signals under the policy's actions. If hallucinated trajectories are directly used for policy optimization, RL is incentivized to exploit systematic model errors rather than true task progress.
This leads to a critical question:
\vspace{-0.6em}
\begin{center}
\emph{If world models inevitably hallucinate, how can RL remain reliable under imperfect imagined dynamics?}
\end{center}
\vspace{-0.6em}

\begin{figure*}[htbp]
    \centering
    \includegraphics[width=\textwidth]{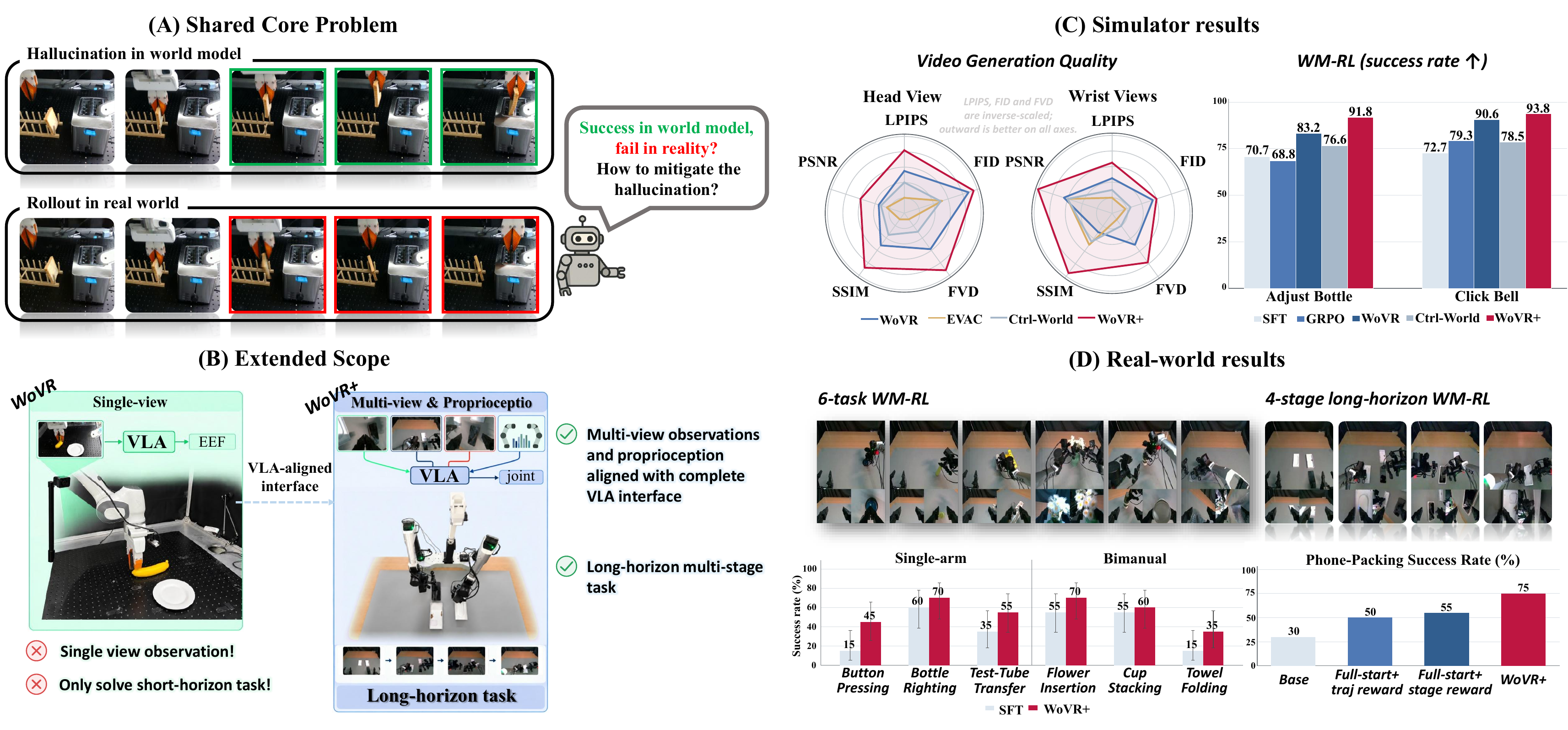}
    \caption{
    \textbf{Overview of motivation, extensions, and evaluation of WoVR+.}
    (A) \textbf{Shared core problem.}
    World-model hallucinations can produce spurious imagined successes that fail to transfer to real execution, motivating hallucination-aware WM-RL.
    (B) \textbf{Extended scope.}
    WoVR+ extends WoVR's reliability-oriented framework to complete VLA interfaces with  multi-view observations and proprioceptive feedback, as well as long-horizon, multi-stage manipulation.
    (C) \textbf{Simulator results.}
    WoVR+ improves multi-view video-generation fidelity and enables effective policy optimization through WM-RL.
    (D) \textbf{Real-world results.}
    The resulting policies improve performance across six real-robot manipulation tasks and support end-to-end optimization of a 4-stage long-horizon task.
    }
    \label{fig:overview}
    \vspace{-4mm}
\end{figure*}

WoVR~\cite{jiang2026wovrworldmodelsreliable} argues that using world models for RL is not primarily a modeling problem, but a reliability problem, and controls hallucination at three interconnected levels: controllable simulator design, reliable imagined interaction, and policy--model alignment. Related efforts have also sought to mitigate the impact of unreliable imagination by down-weighting rewards from hallucinated rollouts~\cite{chen2026hawmpohallucinationawareworldmodelbased}, grounding synthetic experience in real trajectories and regulating its use~\cite{xue2026worldsampleclosedlooprealrobotrl}, and shortening imagined trajectories through branched rollouts initialized from recorded states~\cite{zhang2026practicalworldmodelbasedreinforcement}.

Despite these efforts, extending reliable WM-RL to modern VLA policies and long-horizon tasks introduces additional challenges. 
First, unlike the predominantly single-view ACWMs~\cite{WMPO,jiang2026wovrworldmodelsreliable}, modern VLA policies~\cite{pi05,openvlaoft} rely on multi-view observations and robot proprioception for fine-grained manipulation, requiring world models to maintain temporal, spatial, and cross-modal consistency.
Second, during WM-RL training of flow-based action-chunk policies, overly stochastic action exploration can disrupt action smoothness and exacerbate world-model hallucinations. Maintaining effective exploration while suppressing exploration-induced action jitter is therefore essential to preserving the reliability of imagined interaction.
Finally, world-model generalization errors can accumulate over autoregressive rollouts, limiting the performance of existing WM-RL frameworks on long-horizon, multi-stage tasks, making reliable policy optimization increasingly challenging.

We therefore propose \textbf{WoVR+}, a VLA-Interface-Aligned \textbf{Wo}rld-model-based framework for post-training \textbf{V}ision--Language--Action policies with \textbf{R}einforcement Learning, preserving WoVR's overall WM-RL framework while redesigning its learned environment and imagined interaction protocol (Fig.~\ref{fig:overview}).
We construct a geometry-grounded world model that provides multi-view prediction and proprioceptive feedback, aligning the learned environment with the VLA interface while mitigating hallucinations.
To support reliable imagined interaction under exploration, we adopt chunk-wise noise, promoting meaningful, temporally coherent exploration while suppressing jitter.
For long-horizon tasks, we introduce stage-local imagined training, using short rollouts from stage-entry states and localized supervision to provide direct, balanced training opportunities across stages while avoiding long generated prefixes.

Our contributions are summarized as follows:
\begin{itemize}
    \item \textbf{Extending hallucination-aware WM-RL framework across richer settings.}
    Building on WoVR's central premise that reliable policy optimization requires explicitly controlling hallucination in imagined interaction, we extend the same WM-RL framework to richer VLA interfaces and long-horizon, multi-stage manipulation.
    The present work broadens the reliability problem to include interface completeness, exploration-induced prediction degradation, and stage-wise learning over long horizons.

    \item \textbf{Substantial redesigns for the extended WM-RL setting.}
    Rather than directly reusing WoVR's original components, we redesign the learned environment around geometry-grounded multi-view prediction and proprioceptive feedback to align with VLA interface while mitigating hallucinations;
    introduce chunk-wise Flow-SDE exploration to suppress exploration-induced jitter that can degrade world-model predictions; 
    and develop stage-local imagined training with stage-entry resets, stage-specific rewards, local horizons, and controllable sampling to balance training opportunities across stages while avoiding long generated prefixes for long-horizon tasks.

    \item \textbf{Systematic evaluation in simulation and on real robots.}
    Experiments validate resulting framework from WM prediction to downstream policy improvement. Beyond performance gains, analyses demonstrate the benefits of temporally coherent exploration and stage-local long-horizon training.
    WoVR+ improves average success over SFT by $21.09\%$ on 2 RoboTwin tasks and $16.7\%$ across 6 real-robot tasks, and increases success on 4-stage phone packaging from $30\%$ to $75\%$.
\end{itemize}

This work substantially extends the hallucination-aware WM-RL framework of WoVR~\cite{jiang2026wovrworldmodelsreliable} to align with modern multi-view VLA interfaces, including proprioceptive feedback, and support long-horizon multi-stage manipulation. Section~\ref{sec:problem_formulation} summarizes the shared framework and clarifies the components retained, redesigned, and introduced in the present work.

\section{Related Work}

\subsection{Multi-View World Models}

Recent ACWMs increasingly support closed-loop interaction with multi-view VLAs. Ctrl-World~\cite{CtrlWorld} alternates chunk-wise multi-view prediction with VLA inference, while WEAVER ~\cite{jain2026weaverbetterfasterlonger} combines multi-view latent dynamics, memory, and reward prediction for policy evaluation, synthetic-data augmentation, and test-time planning. GE-Sim2.0~\cite{qiu2026gesim20roadmapcomprehensive} further predicts multi-view observations together with robot proprioception and success signals, and accelerates generation through DMD2~\cite{yin2024improveddistributionmatchingdistillation}. 
Beyond individual architectures, some studies have systematically investigated design choices for ACWMs: GigaWorld-1~\cite{gigaworldteam2026gigaworld1roadmapbuildworld} analyzes pretraining, action representation, memory, and evaluator-oriented post-training; GeniWorld~\cite{gu2026geniworldgeneralizableinteractiveworld} studies URDF-rendered visual action representations. 
Despite these advances, existing multi-view ACWMs mainly emphasize
prediction and evaluation, while their use as closed-loop environments for VLA-RL places stricter requirements on interface consistency across observations, actions, camera motion, and proprioceptive transitions~\cite{chen2026roboticworldmodelsreally}.

\subsection{World-Model-Based VLA Reinforcement Learning}

Recent works directly optimize VLA policies through interactions with ACWMs.
World-Env~\cite{xiao2026worldenvleveragingworldmodel} and WMPO~\cite{WMPO} construct learned environments for VLA policy optimization through pixel-space rollouts, with World-Env building on EVAC's~\cite{jiang2025enerverseacenvisioningembodiedenvironments} world model design. ProphRL~\cite{zhang2026reinforcingactionpoliciesprophesying} extends imagined policy optimization to flow-based action policies~\cite{flowgrpo}.
WoVR~\cite{jiang2026wovrworldmodelsreliable} mitigates hallucination through shortened rollouts and policy-model adaptation and World-VLA-Loop~\cite{liu2026worldvlaloopclosedlooplearningvideo} alternates policy improvement with world-model refinement.
VLA-MBPO~\cite{zhang2026practicalworldmodelbasedreinforcement} further combines a world model with short branched rollouts to reduce compounding prediction errors.
Despite these advances, existing WM-RL methods either do not explicitly address the reliability of world-model-based policy optimization or focus on individual reliability issues in isolation. How to jointly construct a VLA-aligned learned environment, support effective policy exploration, and organize reliable imagined policy optimization remains insufficiently studied.


\section{WoVR framework and Extensions}
\label{sec:problem_formulation}

Building on the same core premise that controlling hallucination enables reliable policy optimization in imagination, we retain WoVR’s overall WM-RL framework and reliability-driven design principles. Within this framework, we systematically redesign the world model, exploration noise, and imagined interaction protocol to support RL post-training of multi-view VLA policies for long-horizon manipulation. This section first reviews WoVR and the shared WM-RL framework, then clarifies the components retained and the extensions introduced in this work.

\subsection{WM-RL framework in WoVR}
\begin{figure}[bthp]
\centering
\includegraphics[width=\linewidth]{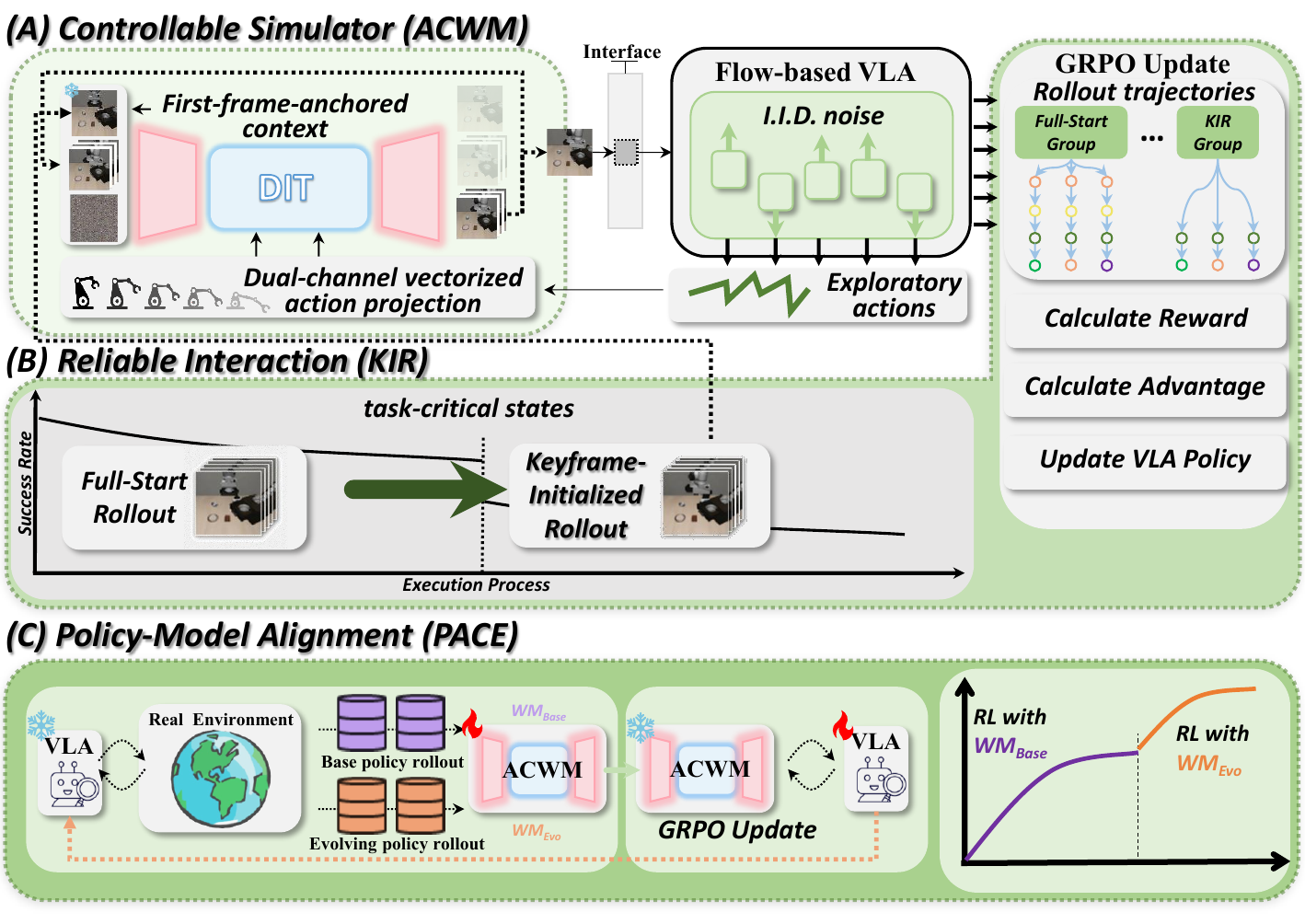}
\caption{\textbf{WM-RL framework in WoVR, redrawn based on~\cite{jiang2026wovrworldmodelsreliable}.} WoVR treats the world model as a generative simulator and controls hallucination across three levels.}
\vspace{-4mm}
\label{fig:wovr}
\end{figure}

WoVR mitigates hallucination in WM-RL through three complementary mechanisms (Fig.~\ref{fig:wovr}): a controllable simulator, reliable interaction, and policy–model alignment.

\subsubsection{Simulator-level control}
WoVR constructs an ACWM by injecting actions into a pretrained video generator through AdaLN and Cross-Attention, while using first-frame-anchored context to stabilize autoregressive rollouts.

Before WM-RL, a base VLA policy $\pi_{\theta_0}$, obtained through SFT on robot demonstrations, is executed in the real environment to collect transition data. The world model $W_\phi$ is then trained on these recorded transitions:
\begin{equation}
\label{eq:wovr_acwm_training}
\begin{aligned}
\mathcal{D}_{\mathrm{ACWM}}
&=
\left\{
\left(
O_{\iota_t}, A_t, O_{t+1:t+H_w}
\right)
\right\}, \\
\phi_0
&=
\arg\min_{\phi}\;
\mathbb{E}_{\mathcal{D}_{\mathrm{ACWM}}}
\left[
\mathcal{L}_{\mathrm{ACWM}}(W_\phi)
\right].
\end{aligned}
\end{equation}
Here, $O_{\iota_t}$ combines the first-frame anchor with recent observations, $A_t$ denotes the executed action sequence, and $O_{t+1:t+H_w}$ provides the recorded future observations over ACWM prediction horizon $H_w$. The loss $\mathcal{L}_{\mathrm{ACWM}}$ is the flow-matching objective for action-conditioned video prediction.

\subsubsection{Interaction-Level Reshaping}
Once trained, the world model is frozen during each
policy-optimization phase and serves as a learned interaction environment. The VLA policy interacts with it in a closed loop to generate groups of imagined rollouts from shared reset contexts. A reward model evaluates these rollouts, and the policy is updated using the clipped GRPO objective:
\begin{equation}
\label{eq:wovr_grpo}
\begin{aligned}
&J_{\mathrm{GRPO}}(\theta) 
= \mathbb{E}\Bigg[
\frac{1}{G}\sum_{i=1}^{G}
\frac{1}{T_i^{\mathrm{valid}}}
\sum_{t=1}^{T_i^{\mathrm{valid}}}
\min\!\Big(
\rho_t^{(i)}(\theta)\hat{A}^{(i)}, \\
&\qquad\quad
\operatorname{clip}\!\big(
\rho_t^{(i)}(\theta),\,1-\epsilon,\,1+\epsilon
\big)\hat{A}^{(i)}
\Big)
\Bigg].
\end{aligned}
\end{equation}
The valid length $T_i^{\mathrm{valid}}$ extends to the first detected success, or to the rollout horizon if no success is detected; subsequent steps are masked from optimization.

This objective complements Keyframe-Initialized Rollout (KIR), which initializes a portion of imagined rollouts from recorded contexts near task-critical intermediate states.
By avoiding long generated prefixes, KIR reduces the effective prediction depth and the associated accumulation of model errors. Trajectory-length normalization assigns greater per-timestep weight to shorter rollouts, allowing their task-critical segments to contribute.

\begin{figure*}[!t]
\centering
\includegraphics[width=\linewidth]{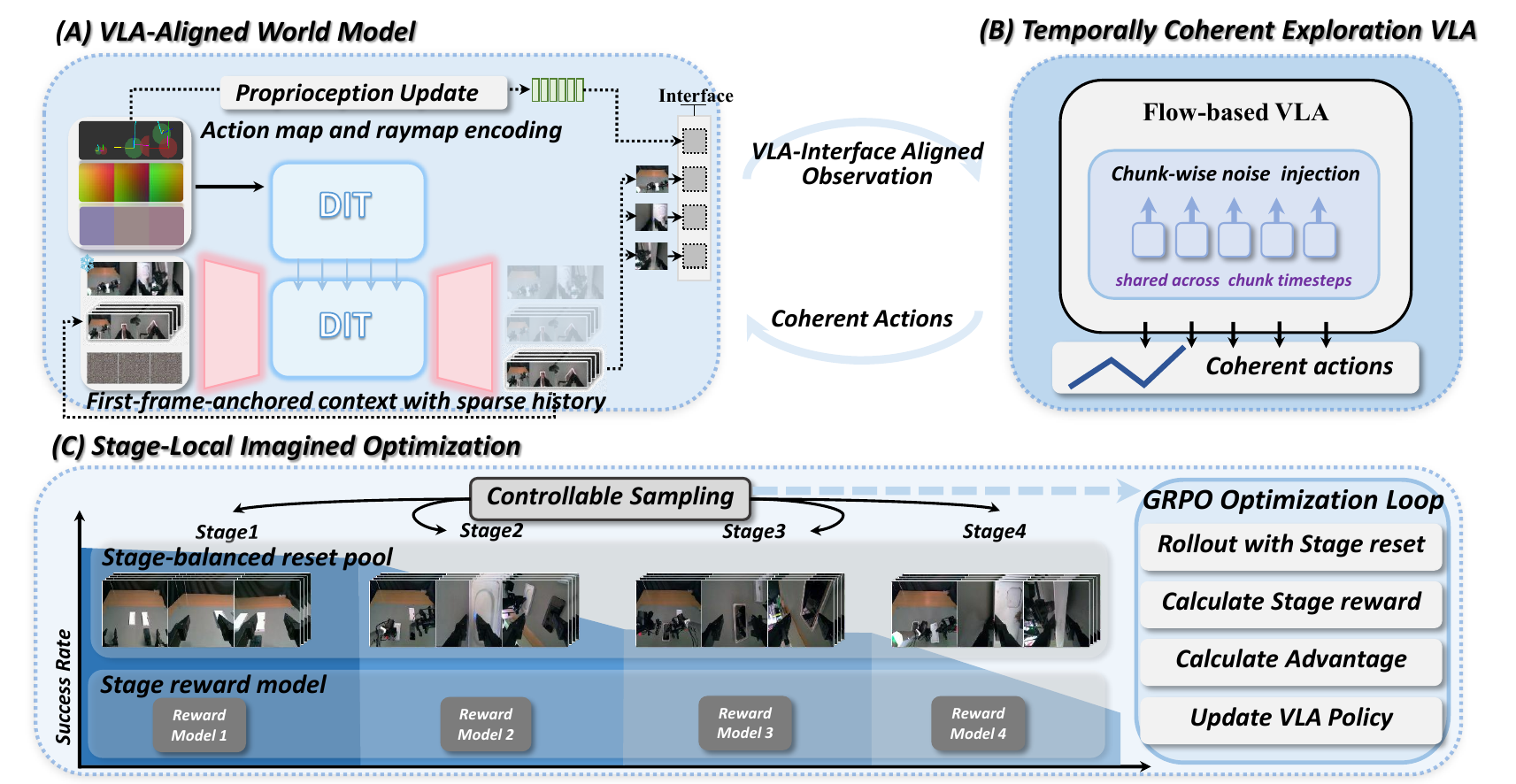}
\caption{
\textbf{Overview of WoVR+.}
WoVR+ extends the reliability-oriented framework with a VLA-aligned multi-view world model with proprioceptive feedback, chunk-wise Flow-SDE exploration and stage-balanced reset-based local rollouts for long-horizon policy optimization.
}
\vspace{-4mm}
\label{fig:wovr_vs_iwovr}
\end{figure*}

\subsubsection{Alignment-Level Regulation}
As WM-RL progresses, the policy's action distribution can
drift away from the distribution represented in the world-model training data, degrading the reliability of imagined transitions and aggravating hallucination. 
WoVR addresses this mismatch through Policy-Aligned Co-Evolution (PACE). After an initial phase of imagined policy optimization, the evolved VLA policy is executed in the real environment to collect a limited set of additional trajectories.
These trajectories are combined with the original base-policy rollouts to further fine-tune the world model using the same objective.
The refined world model is then frozen, and policy optimization resumes through imagined interaction within the updated environment. 
This low-frequency recalibration mitigates policy--model mismatch without requiring continuous real-environment interaction or world-model updates during each policy-optimization phase.

\subsection{Extensions and Scope of the Present Work}
Building on the WM-RL framework described above, WoVR+ extend WoVR's hallucination-aware design to support multi-view VLA policies and long-horizon, multi-stage manipulation (Fig.~\ref{fig:wovr_vs_iwovr}).

The following summarizes the components retained, redesigned, or newly introduced, and clarifies the scope of the present study:
\begin{itemize}

    \item \textbf{Extended and Redesigned.}
    We retain WoVR's first-frame anchoring principle and extend its temporal context to an anchored sparse history that combines the persistent scene anchor with fixed-stride recent observations. We redesign the interaction interface and action-conditioning mechanism.
    The single-view, delta-EEF formulation with vector action conditioning in WoVR is replaced by a VLA-aligned multi-view interface with absolute joint-and-gripper commands, geometry-grounded action and camera conditioning, and command-based proprioceptive updates.

    \item \textbf{New.}
    We introduce chunk-wise Flow-SDE exploration for absolute-action chunks, preserving the marginal exploration scale while enforcing temporal coherence across neighboring action targets.

    \item \textbf{Broadened in scope.}
    KIR in WoVR uses intermediate-state initialization around a task-critical region to reduce effective prediction depth.
    In the present work, we carry this reset-based principle into long-horizon multi-stage optimization, combining data-supported stage-entry resets with stage-specific rewards, local rollout horizons, and controllable stage sampling to organize training opportunities across task stages.

\end{itemize}
\section{Methods}
\label{method}

As summarized in Fig.~\ref{fig:wovr_vs_iwovr}, WoVR+
consists of three coupled components:
a \textbf{VLA-Aligned World Model} (Sec.~\ref{sec:reliable_world_model}),
\textbf{Temporally Coherent Exploration} (Sec.~\ref{sec:action_exploration}), and
\textbf{Stage-Local Imagined Optimization} (Sec.~\ref{sec:stage_training}).


\subsection{A VLA-Aligned Multi-View World Model for Reliable Local Dynamics}
\label{sec:reliable_world_model}

\begin{figure}[bthp]
\centering
\includegraphics[width=\linewidth]{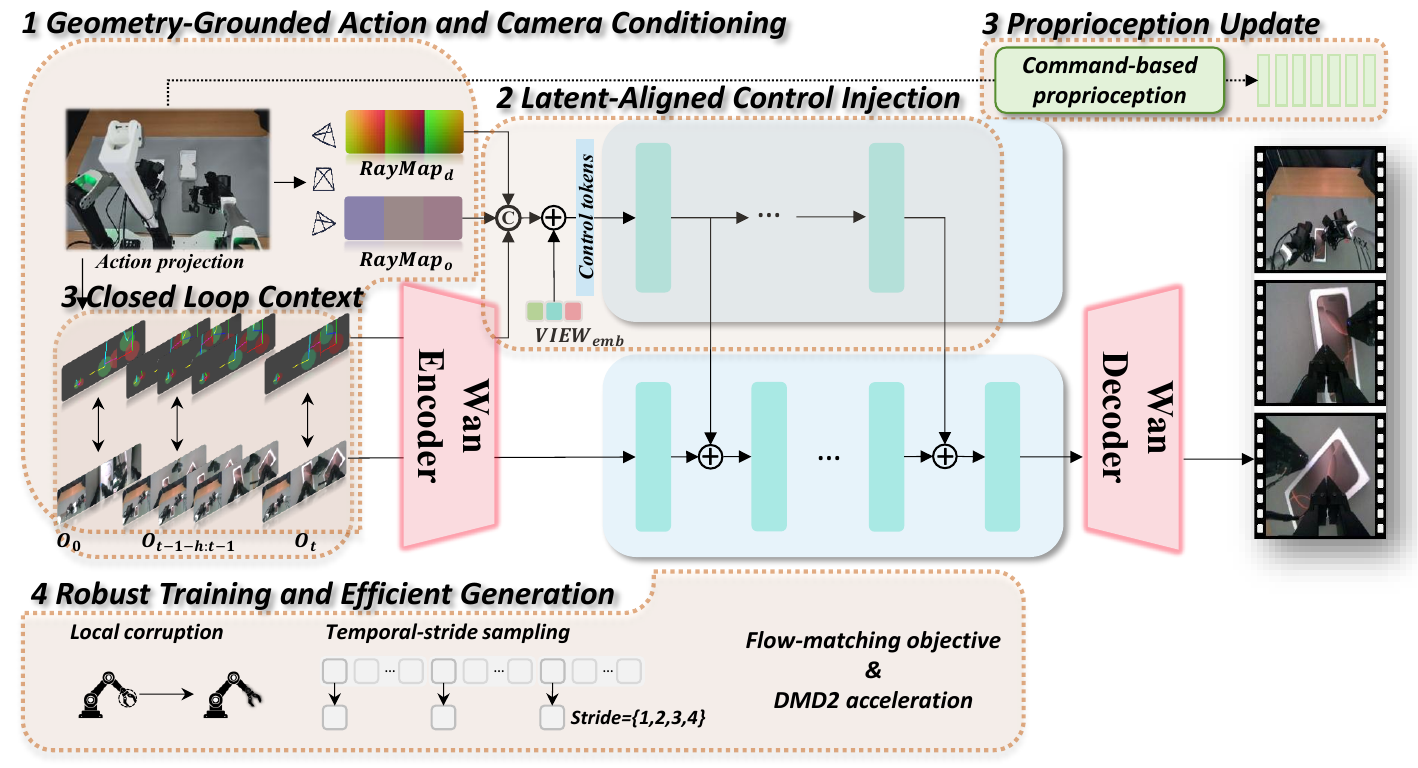}
\caption{\textbf{Architecture of the VLA-aligned world model.}
Action maps and camera ray maps are explicitly encoded as geometric conditions and injected into the backbone through a parallel DiT control branch, enabling multi-view video prediction.}
\label{fig:worldmodelstructure}
\end{figure}

We instantiate the visual transition model as \(W_{\phi}\). For WM-RL, reliability requires more than visual fidelity: the model must respond to robot actions faithfully and provide observations and proprioception needed for subsequent policy decisions. We therefore construct a multi-view model with explicit action and camera geometry, an anchored temporal
context, and a command-based proprioceptive update. Given an absolute-joint action chunk $A_{t}$, the visual transition is
\begin{equation} \widehat{O}_{t+1:t+H_{w}} \sim W_{\phi} \left( \cdot \mid O_{\iota_{t}}, A_{t}; \mathcal{K}, \mathcal{E} \right), \label{eq:multiview_world_transition} \end{equation} 
where \(O_{\iota_{t}}\) is the sparsely sampled multi-view history, \(\mathcal{K}\) and \(\mathcal{E}\) denote camera intrinsics and extrinsics. $H_{w}$ is the prediction horizon of one ACWM call. Fig~\ref{fig:worldmodelstructure} summarizes the ACWM, which uses Wan2.2-TI2V-5B~\cite{wan2025wanopenadvancedlargescale} as the video backbone and a VACE parallel control branch~\cite{jiang2025vaceallinonevideocreation} for geometric conditioning.

\subsubsection{Geometry-Grounded Action and Camera Conditioning}
\paragraph{Spatial action maps}
WoVR injects action conditions through Cross-Attention and AdaLN, which provides action conditioning but does not explicitly encode the view-dependent geometric consequences of robot motion.
To support  prediction across head and wrist views and improve responsiveness to action perturbations, we replace this vector conditioning with view-specific geometric control.
To establish an explicit correspondence between joint commands and image-space motion, we convert absolute-joint targets into end-effector (EEF) poses through forward kinematics and project them into view-specific action maps~\cite{jiang2025enerverseacenvisioningembodiedenvironments,zhang2026reinforcingactionpoliciesprophesying,chen2026bridgev2wbridgingvideogeneration}. EEF is represented by a circle and three projected local coordinate axes: the axes encode orientation, while the center of the circle represents the projection position of EEF, the diameter encodes the projection depth, and the color of the circle encodes the degree of the gripper's opening.
Given arm $b$ and target time $t$, EEF pose in camera $v$ is
\begin{equation}
{}^{c_v}\mathbf T_{\tilde e,t}^{(b)}
=
\left({}^{w}\mathbf T_{c_v,t}\right)^{-1}
\operatorname{FK}_{b}\!\left(\mathbf q_{t}^{(b)}\right)
{}^{e}\mathbf T_{\tilde e}.
\label{eq:action_map_pose}
\end{equation}
where $q_{t}^{(b)}$ is the absolute-joint target, ${}^{w}\mathbf T_{c_{v},t}$ is the camera-to-world pose and ${}^{e}\mathbf T_{\tilde e}$ is a fixed local visualization offset. The projected circle radius is computed as
\begin{equation}
\rho_{v,t}^{(b)}
=
\operatorname{clip}\!\left(
\rho_{\mathrm{ref}}
\frac{z_{\mathrm{ref}}}{z_{v,t}^{(b)}},
\rho_{\min},
\rho_{\max}
\right),
\label{eq:depth_scaled_action_radius}
\end{equation}
where $\rho_{\mathrm{ref}}$ is the pixel radius at reference depth
$z_{\mathrm{ref}}$. Its color
$\mathbf c_b(\hat g_{t}^{(b)})$ encodes the normalized gripper value.

\begin{figure}[htbp]
\centering
\includegraphics[width=0.8\linewidth]{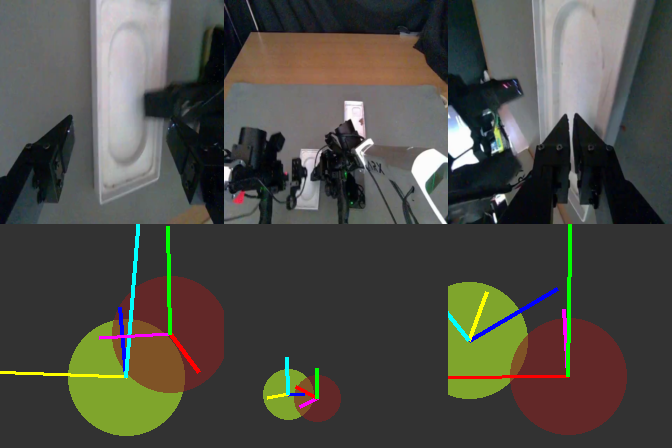}
\caption{\textbf{Construction of Action Map in Fig.~\ref{fig:worldmodelstructure}.} Projected EEF axes encode orientation, circle color indicates gripper openness, and depth-dependent circle
size supplies perspective information. Semi-transparent compositing retains overlapping bimanual control cues.}
\label{fig:actionmap}
\end{figure}

We adapt this representation to multi-view bimanual interaction by compositing the circles semi-transparently. The resulting maps retain perspective cues and overlapping signals from multiple
arms, as illustrated in Fig~\ref{fig:actionmap}. Repeating this construction
along the action horizon gives one action-map sequence per
view. The sequences are arranged identically to the multi-view
observations before VAE encoding. 

\paragraph{Camera Raymaps}
Action map specifies where EEF projects, but the head and wrist views differ in camera pose and motion. We therefore encode per-view ray origins and directions as continuous geometric conditions~\cite{jiang2025enerverseacenvisioningembodiedenvironments,qiu2026gesim20roadmapcomprehensive}.

Rather than rasterizing signed ray coordinates as quantized
RGB images, we construct them directly on the spatial grid of
the video latent, avoiding an additional VAE encoding path. For
a latent-grid location $(x,y)$, let the camera-to-world transform
be $(R_{v,u},\mathbf{t}_{v,u})$ and let $K_v^z$ be the intrinsic matrix scaled to the latent resolution. The corresponding ray is
\begin{equation}
\begin{aligned}
    \mathbf{o}_{v,u}^z(x,y)&=\mathbf{t}_{v,u},\\
    \mathbf{d}_{v,u}^z(x,y)&=
    \frac{R_{v,u}(K_v^z)^{-1}\overline{\mathbf{p}}(x,y)}
    {\left\|R_{v,u}(K_v^z)^{-1}\overline{\mathbf{p}}(x,y)\right\|_2},
\end{aligned}
\label{eq:wm_camera_rays}
\end{equation}
where $\overline{\mathbf{p}}(x,y)=[x+\tfrac12,y+\tfrac12,1]^\top$.
Camera poses are updated at each frame using forward kinematics and fixed hand--eye calibration.
Thus, the geometric condition describes both the commanded manipulation and the camera configuration from which it is observed.

\subsubsection{Latent-Aligned Control Injection}
The action map and raymap representations must align with the video latent in both time and space.
Let $Z^{\mathrm{act}}=E_{\mathrm{VAE}}(\mathcal{M})$ denote the latent of the assembled action-map sequence $\mathcal{M}$.
The Wan VAE retains a separate first-frame latent slot and compresses subsequent frames in groups of four.
We accordingly associate each latent slot with four ray origins and four ray directions, concatenating their coordinates along the channel dimension.
The first-frame rays are repeated four times for the initial slot.
Per-view tensors are then assembled using the same spatial layout as the video and action maps.

\begin{figure}[htbp]
    \centering
    \includegraphics[width=\linewidth]{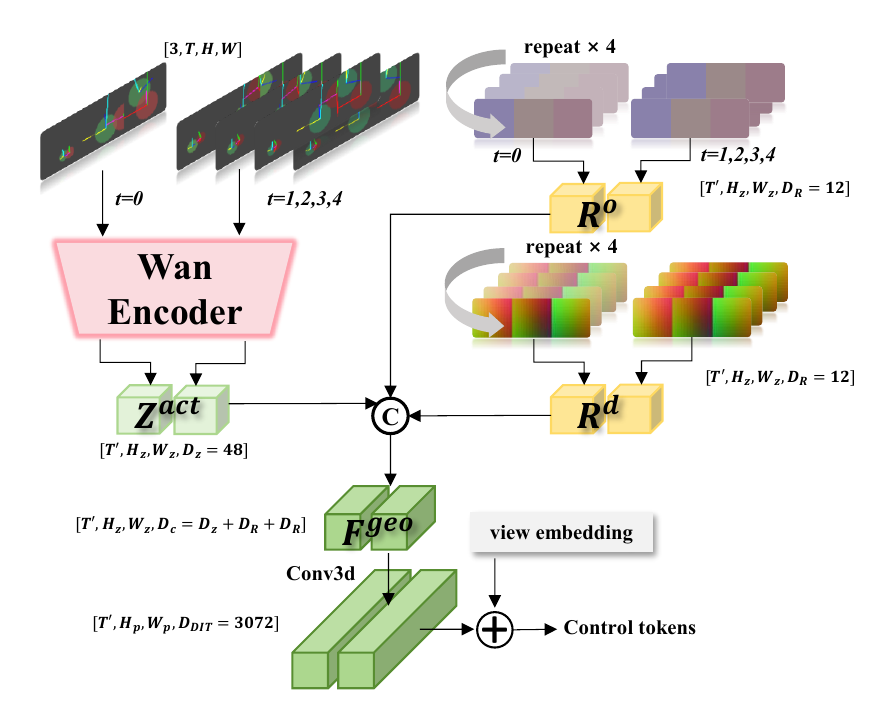}
    \caption{\textbf{Construction of latent-aligned geometric condition in Fig.~\ref{fig:worldmodelstructure}.} The VAE-encoded action map is concatenated with temporally grouped ray origins and directions, projected into the VACE hidden space, and augmented with view embeddings. $H_z,W_z$ denote the assembled video-latent size; the patch projection produces $H_p,W_p$.}
    \label{fig:wm_geometry_condition}
\end{figure}

Denote the resulting tensors by $R^o$ and $R^d$.
They are fused with the action latent and projected to the control hidden space:
\begin{equation}
\begin{aligned}
    F^{\mathrm{geo}}&=\operatorname{Concat}_{\mathrm{ch}}
        \!\left(Z^{\mathrm{act}},R^o,R^d\right),\\
    g_{\ell,x,y}&=
        \left[\mathcal{P}\!\left(F^{\mathrm{geo}}\right)\right]_{\ell,x,y}
        +\mathbf{e}^{\,v(x,y)},
\end{aligned}
\label{eq:wm_geometry_fusion}
\end{equation}
where $\mathcal{P}$ is a 3D patch embedding and $\mathbf{e}^{\,v(x,y)}$ is a learned embedding identifying the camera view occupying the corresponding spatial region.
Fig.~\ref{fig:wm_geometry_condition} illustrates this construction.

We inject this view-aware condition through a parallel DiT control stream rather than replacing the pretrained video backbone.
Let $\mathcal{B}_j$ and $\mathcal{C}_j$ denote the $j$th backbone and control blocks, respectively.
Their features are updated as
\begin{equation}
\begin{aligned}
    c_{j+1}&=\mathcal{C}_j(c_j,g),\\
    h_{j+1}&=\mathcal{B}_j(h_j)
        +\mathbf{1}[j\in\mathcal{S}]\,P_j(c_{j+1}),
\end{aligned}
\label{eq:wm_control_injection}
\end{equation}
where $\mathcal{S}$ specifies the backbone blocks with control injection and $P_j$ is a learnable projection.
The control branch learns the incremental effects of robot actions and camera geometry while retaining the backbone architecture and its pretrained initialization.
Both the main DiT and the control branch are trainable; the VAE and text encoder remain frozen.

\subsubsection{Closed Loop Context and Proprioception Update}
\paragraph{Anchored sparse history}
During autoregressive interaction, generated observations become the context for subsequent ACWM calls.
Recent observations describe task progress but may gradually lose initial scene information; conversely, the first frame alone does not describe the current state. Therefore, based on WoVR, we combine a persistent first-frame anchor with fixed-stride recent history~\cite{jiang2026wovrworldmodelsreliable,qiu2026gesim20roadmapcomprehensive}:
\begin{equation}
    \iota_t=\left(0,[t-(C-2)s]_+,\ldots,[t-s]_+,t\right),
    \label{eq:wm_sparse_history}
\end{equation}
where $[x]_+=\max(0,x)$, $C=5$ is the context length, and $s=4$ is the history stride.
Clipping early indices to zero supplies a valid context near initialization, while the fixed context budget retains both the scene anchor and recent task information.

\paragraph{Command-based proprioception}
The policy also requires the joint and gripper state at its proprioception input.
Under the absolute-position interface, we approximate this sequence by shifting the commanded targets by one step:
\begin{equation}
    \widehat{s}_{t+1:t+H}=a^{\mathrm{joint}}_{t:t+H-1},
    \label{eq:wm_proprioception}
\end{equation}
where $a^{\mathrm{joint}}$ includes absolute joint and gripper targets.
This command-following approximation avoids an additional learned proprioception model. 
This is enabled by the absolute-joint interface: unlike delta-EEF commands, the policy output directly specifies the joint target required by the next VLA query. We evaluate this approximation with learned transition model in Appendix~\ref{app:command_proprio}. 

\subsubsection{Robust Training and Efficient Generation}
\paragraph{Training for recursive prediction}
The ACWM is trained on real observations but receives its own generated observations during WM-RL.
To reduce this context gap, we apply geometry-aware local corruption to manipulation-relevant regions of non-anchor context frames.
A 3D region near a selected EEF is projected into the context frames of one selected camera and blurred, while the first-frame anchor and all other views remain unchanged.
We also sample temporal strides from $\{1,2,3,4\}$, synchronizing the selected observations, actions, action maps, and camera conditions.
This exposes the model to different motion magnitudes and prediction intervals without collecting additional trajectories.

\paragraph{Flow-matching objective and inference}
Let $z_0$ denote the clean target video latent and let $\varepsilon\sim\mathcal{N}(0,I)$.
We form $z_\tau=(1-\tau)z_0+\tau\varepsilon$ and train a velocity predictor conditioned on the sparse history and geometric control:
\begin{equation}
    \mathcal{L}_{\mathrm{FM}}=
    \mathbb{E}\!\left[
    \left\|v_\phi(z_\tau,\tau\mid O_{\iota_t},g)
    -(\varepsilon-z_0)\right\|_2^2\right].
    \label{eq:wm_flow_matching}
\end{equation}
We use five integration steps for the standard ACWM and further apply DMD2~\cite{yin2024improveddistributionmatchingdistillation} to obtain a one-step generator for higher-throughput imagined interaction.

\subsection{WM-RL with Temporally Coherent Exploration}
\label{sec:action_exploration}

We optimize flow-based VLA policies in the learned environment
using the Flow-SDE sampling formulation of
$\boldsymbol{\pi}_{\mathbf{RL}}$~\cite{chen2026pitextttrlonlinerlfinetuning} and GRPO~\cite{deepseekmath}.
We first describe the stochastic sampling procedure, then
introduce chunk-wise noise for temporally coherent exploration.

\subsubsection{From Flow-ODE to Flow-SDE}
\label{sec:flow_sde_sampling}

For one policy query, let $\mathbf{c}$ denote the current
multi-view observations, proprioception, and task instruction.
A flow-based policy generates an action chunk by transporting
Gaussian initial noise through an ordinary differential
equation (ODE):
\begin{equation}
    \frac{\mathrm{d}\mathbf{x}^{\tau}}{\mathrm{d}\tau}
    =
    \mathbf{v}_{\theta}
    (\mathbf{x}^{\tau},\tau \mid \mathbf{c}),
    \label{eq:policy_flow_ode}
\end{equation}
where $\mathbf{x}^{\tau}\in\mathbb{R}^{H_a\times D}$ is the
intermediate action chunk, $H_a$ is the action horizon, and
$D$ is the action dimension.
The integration is deterministic given the initial noise
and conditioning input.

To introduce stochastic transitions with tractable conditional
likelihoods, we adopt the ODE-to-SDE conversion of
$\pi_{\mathrm{RL}}$~\cite{chen2026pitextttrlonlinerlfinetuning}.
At a selected integration step, the discretized update is
\begin{equation}
\begin{aligned}
    \mathbf{x}_{k+1,h}
    &=
    \boldsymbol{\mu}_{\theta,k,h}
    + \sigma_k \boldsymbol{\epsilon}^{\mathrm{iid}}_{k,h},
    \\
    \boldsymbol{\epsilon}^{\mathrm{iid}}_{k,h}
    &\overset{\mathrm{iid}}{\sim}
    \mathcal{N}(\mathbf{0},\mathbf{I}_D),
\end{aligned}
\label{eq:iid_flow_sde}
\end{equation}
where $k$ indexes flow integration steps and
$h\in\{1,\ldots,H_a\}$ indexes temporal positions within the
action chunk.
Here, $\boldsymbol{\mu}_{\theta,k,h}$ is the drift-corrected
Flow-SDE transition mean, rather than the original ODE update,
and $\sigma_k$ is the discretized noise standard deviation,
including the integration-step scaling.
These Gaussian transitions provide tractable scores for
policy optimization.
During training, we use one randomly selected SDE step per
action chunk and ODE updates at the remaining steps.

\subsubsection{Chunk-Wise Exploration Noise}
\label{sec:chunkwise_exploration}

\begin{figure}[bthp]
\centering
\includegraphics[width=\linewidth]{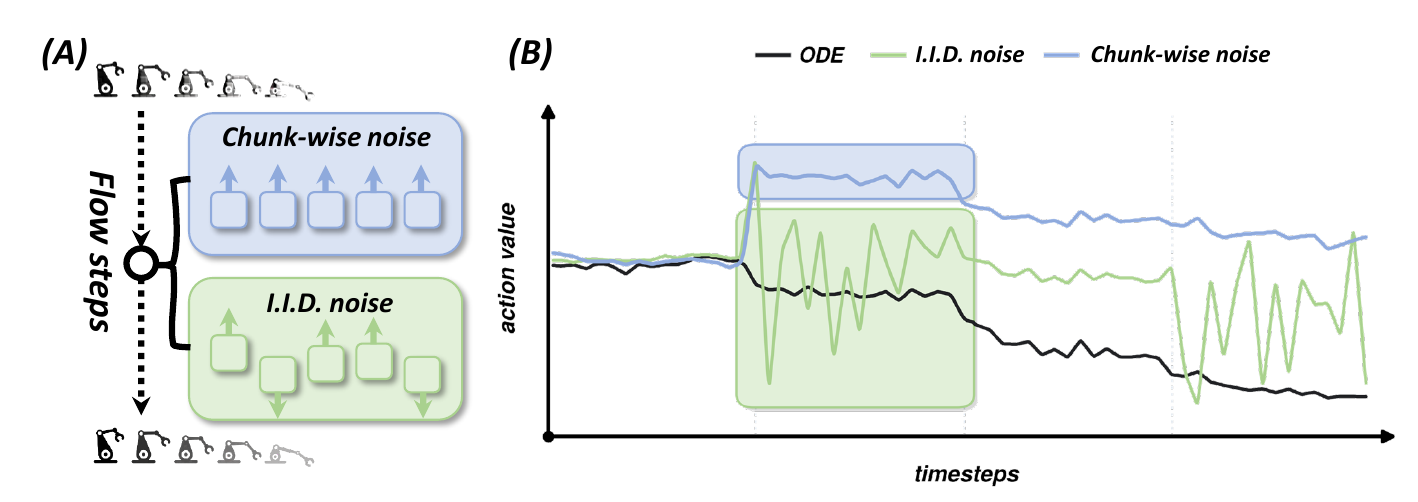}
\caption{
\textbf{Action-space exploration for flow-based action-chunk policies.}
\textbf{(A)} I.I.D. SDE applies independent noise to each action token, whereas chunk-wise SDE shares one noise vector across the horizon.
\textbf{(B)} I.I.D. noise causes high-frequency action fluctuations, while chunk-wise noise preserves per-token exploration with temporally coherent deviations from ODE.
}

\label{fig:chunkwisenoise}
\end{figure}
Unlike the delta-EEF interface used in WoVR, each token in our absolute-joint action chunk directly specifies a target robot configuration.
Consequently, applying independent Flow-SDE innovations to neighboring tokens perturbs consecutive targets in different directions, producing high-frequency variations along the predicted action sequence.

Simply reducing the noise scale suppresses these fluctuations, but at the cost of weaker exploratory perturbations and substantially larger gradient magnitudes, which can destabilize policy optimization.

We instead preserve the marginal innovation scale at each
action token while reshaping its temporal covariance.
At the selected Flow-SDE step, we replace the I.I.D. innovation
in Equation~\ref{eq:iid_flow_sde} with
\begin{equation}
\begin{aligned}
    \boldsymbol{\epsilon}^{\mathrm{cw}}_{k,h}
    &= \mathbf{z}_k,
    \qquad
    \mathbf{z}_k
    \sim \mathcal{N}(\mathbf{0},\mathbf{I}_D),
    \\[2pt]
    \operatorname{Cov}\!\left(
        \boldsymbol{\epsilon}^{\mathrm{iid}}_{k,h},
        \boldsymbol{\epsilon}^{\mathrm{iid}}_{k,h'}
    \right)
    &= \mathbf{0},
    \qquad h \neq h',
    \\[2pt]
    \operatorname{Cov}\!\left(
        \boldsymbol{\epsilon}^{\mathrm{cw}}_{k,h},
        \boldsymbol{\epsilon}^{\mathrm{cw}}_{k,h'}
    \right)
    &= \mathbf{I}_D,
    \qquad h \neq h'.
\end{aligned}
\label{eq:chunkwise_flow_sde}
\end{equation}
Noise sharing is restricted to temporal positions within
the sampled action chunk.
The initial Gaussian noise, transition mean, and noise
schedule remain unchanged.

Thus, each temporal position retains the same Gaussian
innovation marginal, while neighboring action tokens receive
a coherent perturbation rather than independent deviations.
Chunk-wise noise therefore reallocates exploration toward
temporally coherent directions instead of merely reducing
its per-token magnitude, as illustrated in
Fig.~\ref{fig:chunkwisenoise}.

The sampled action chunks drive closed-loop interaction
with the world model, and grouped imagined returns provide
the advantages for GRPO.
We retain the per-token Gaussian transition log-scores to
construct surrogate policy ratios.
Because chunk-wise noise changes the cross-token dependence,
these ratios are not exact joint likelihood ratios of the
correlated action-chunk transition.

\subsection{Stage-Local Imagination for Long-Horizon Manipulation}
\label{sec:stage_training}

\begin{figure}[bthp]
\centering
\vspace{-4mm}
\includegraphics[width=\linewidth]{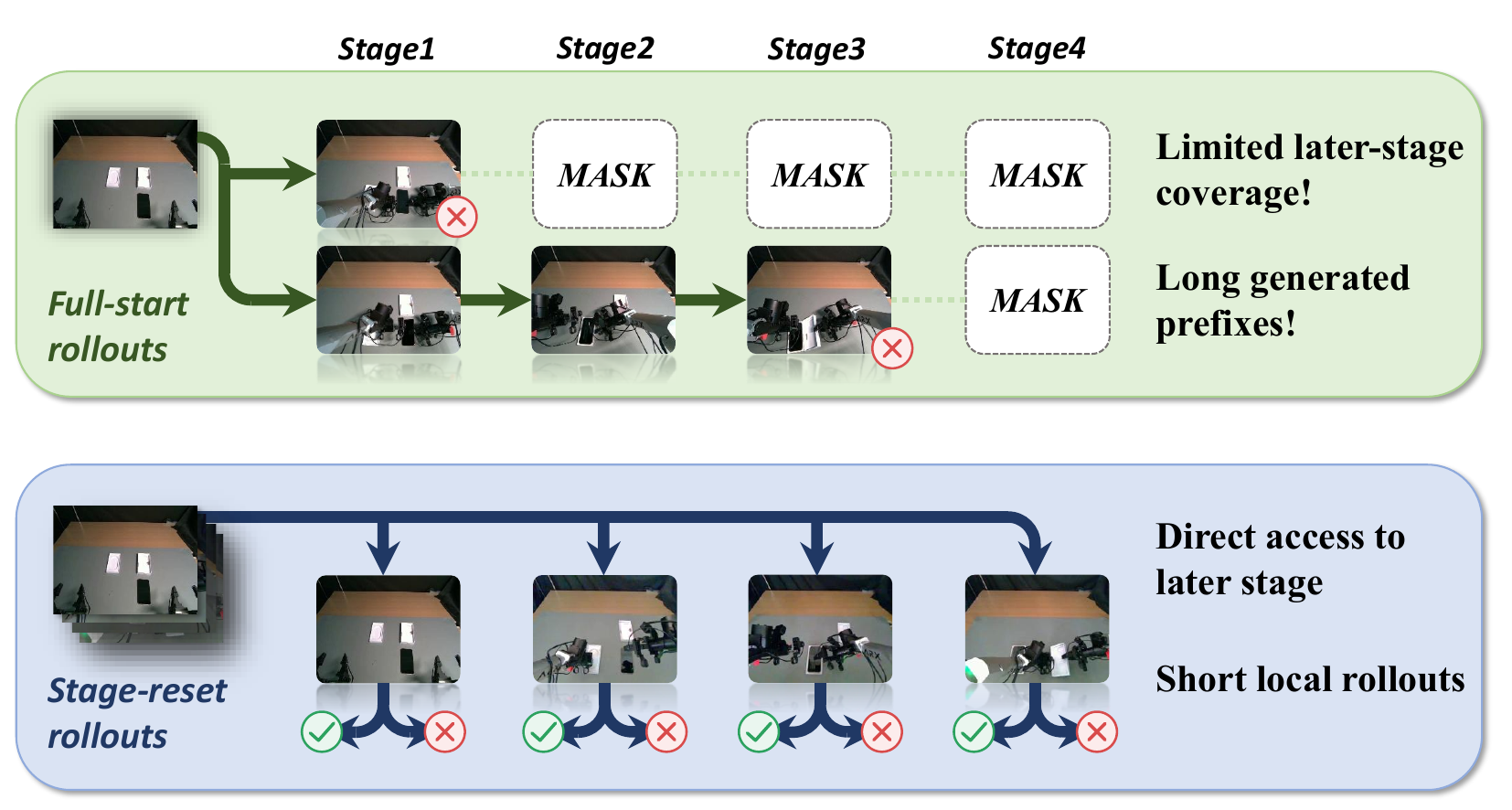}
\caption{\textbf{Stage-balanced imagined interaction.} Stage resets directly initialize
GRPO rollout groups from data-supported stage-entry states, improving
late-stage coverage while avoiding long model-generated prefixes.
}
\label{fig:resetwmrl}
\end{figure}

\subsubsection{Limitations of Full-Start WM-RL}
Long-horizon WM-RL using GRPO with full-start rollouts faces three coupled limitations: insufficient coverage of later-stage states, sparse optimization signals from trajectory-level binary rewards, and compounding world-model errors over long autoregressive rollouts.

Consider a task comprising $M$ causally ordered stages
$\{\mathcal{G}_m\}_{m=1}^{M}$, where $p_m^\pi$ is the probability
that policy $\pi$ completes stage $m$ after reaching it, and
$L_m$ is its local horizon.
Under full-task imagination, the probability of reaching stage
$m$ and the world-model-generated prefix preceding it are
\begin{equation}
    q_m^\pi = \prod_{j=1}^{m-1} p_j^\pi,
    \qquad
    T_m^{\mathrm{pre}} = \sum_{j=1}^{m-1} L_j.
    \label{eq:full_stage_access}
\end{equation}
Consequently, among $N$ full-task rollouts, only $Nq_m^\pi$
are expected to reach stage $m$, causing later-stage training
opportunities to decay multiplicatively.

Let \(T_i^{\mathrm{valid}}\) denote the number of valid rollout steps,
defined as the first-success timestep for successful trajectories and the rollout horizon otherwise, with frames after success or beyond the horizon masked.
Under trajectory-level binary supervision, the reward is assigned only
at the terminal valid step:
\[
r_t^{(i)}
=
y_i\,\mathbf{1}\!\left[t=T_i^{\mathrm{valid}}\right],
\qquad
y_i\in\{0,1\},
\]
where \(y_i\) indicates full-task success.
For a group of $G$ rollouts from the same initial state,
the clipped GRPO objective is
\begin{equation}
\begin{aligned}
    J_{\mathrm{GRPO}}(\theta)
    = \mathbb{E}\Bigg[
    &\frac{1}{G}\sum_{i=1}^{G}
    \frac{1}{T_i^{\mathrm{valid}}}
    \sum_{t=1}^{T_i^{\mathrm{valid}}}
    \min\Big(
        \rho_t^{(i)}(\theta)\hat{A}^{(i)},
        \\
    &\quad
        \operatorname{clip}\!\big(
            \rho_t^{(i)}(\theta),1-\epsilon,1+\epsilon
        \big)\hat{A}^{(i)}
    \Big)
    \Bigg],
\end{aligned}
\label{eq:full_grpo_objective}
\end{equation}
where $\rho_t^{(i)}(\theta)$ is the surrogate policy ratio,
$\epsilon$ is the clipping threshold, and $\hat{A}^{(i)}$
is obtained by normalizing the trajectory return
$R^{(i)}=\sum_t r_t^{(i)}=y_i$ within the group.
Thus, all valid steps share the same outcome-based advantage,
while failed rollouts receive the same return regardless of
their intermediate progress.

Moreover, the rollouts that do reach later stages are
conditioned on increasingly long generated prefixes $T_{i}^{valid}$, making
their imagined states more susceptible to accumulated
world-model errors.

\subsubsection{KIR and Its Limitations}
WoVR~\cite{jiang2026wovrworldmodelsreliable} introduces
Keyframe-Initialized Rollouts (KIR), initializing a portion of WM--VLA rollouts from task-critical states to reduce prediction errors accumulated over long generated prefixes.
Its implementation selects a single task-level
keyframe near the earliest common failure region.

For the keyframe-initialized component, let $k$ denote the
reset index and $T_i^{\mathrm{KIR}}$ the valid length of the
generated suffix.
Using the notation above, its GRPO contribution can be
written as
\begin{equation}
\begin{aligned}
    J_{\mathrm{KIR}}(\theta)
    = \mathbb{E}\Bigg[
    &\frac{1}{G}\sum_{i=1}^{G}
    \frac{1}{T_i^{\mathrm{KIR}}}
    \sum_{t=k+1}^{k+T_i^{\mathrm{KIR}}}
    \min\Big(
        \rho_t^{(i)}(\theta)\hat{A}_{\mathrm{KIR}}^{(i)},
        \\
    &\quad
        \operatorname{clip}\!\big(
            \rho_t^{(i)}(\theta),1-\epsilon,1+\epsilon
        \big)\hat{A}_{\mathrm{KIR}}^{(i)}
    \Big)
    \Bigg],
\end{aligned}
\label{eq:kir_grpo_objective}
\end{equation}
where the expectation is over keyframe contexts and grouped
suffix rollouts.

KIR therefore establishes the usefulness of intermediate-state initialization for reducing effective prediction depth. However, it treats initialization primarily as a rollout-shortening mechanism around a task-level critical state, rather than as a way to organize policy learning across different task stages.

\subsubsection{Stage-Local Imagined Policy Optimization}
We generalize this intermediate-state initialization principle into a stage-organized training protocol, using intermediate-state resets not only to shorten generated prefixes but also to reorganize long-horizon policy optimization.
Instead of selecting a single task-level critical region, we construct data-supported entry distributions for all task stages and couple them with stage-specific rewards, local horizons, and controllable stage-sampling probabilities, thereby jointly changing the rollout initialization distribution, reward decomposition, rollout horizon, and training allocation~\cite{xu2025starevlaprogressivestageawarereinforcement,zhou2026temporalgrpotrajectorylevelcredit}.
To instantiate this stage-organized protocol, we define stage boundaries and entry distributions based on transitions between manipulation skills and observable subgoal completion.
Each stage is assigned a separate binary completion reward, providing localized optimization signals without requiring successful completion of the entire task.

We sample a stage $m_{\mathrm{start}}$ with controllable
weights $\{\omega_m\}_{m=1}^{M}$, and then sample an initial
state $s_0 \sim \rho_{m_{\mathrm{start}}}$ together with its
recorded observation history.
This changes the stage-start probability and generated
prefix to
\begin{equation}
    q_{m,\mathrm{reset}}^{\pi} = \omega_m,
    \qquad
    T_{m,\mathrm{reset}}^{\mathrm{pre}} = 0.
    \label{eq:stage_stage_access}
\end{equation}
We use balanced weights $\omega_m=1/M$, providing each
stage with $N/M$ direct rollout starts in expectation
without replaying any preceding stage.
We then branch a GRPO rollout group from each reset state
for the corresponding local horizon $L_m$, as illustrated
in Fig.~\ref{fig:resetwmrl}.

Let $T_{i,m}^{\mathrm{valid}} \leq L_m$ denote the valid
length of local rollout $i$ for stage $m$, with time
indexed from the reset state.
Only successful completion of the selected stage receives
a nonzero reward:
$r_{t,m}^{(i)}
= y_m^{(i)}\mathbf{1}[t=T_{i,m}^{\mathrm{valid}}]$,
where $y_m^{(i)} \in \{0,1\}$ indicates stage completion,
rather than full-task success.
Frames after detected stage completion or beyond the local
horizon are masked.
The stage-balanced clipped GRPO objective is
\begin{equation}
\begin{aligned}
    J_{\mathrm{stage}}(\theta)
    ={}&
    \sum_{m=1}^{M}\omega_m\,
    \mathbb{E}\Bigg[
    \frac{1}{G}\sum_{i=1}^{G}
    \frac{1}{T_{i,m}^{\mathrm{valid}}}
    \sum_{t=1}^{T_{i,m}^{\mathrm{valid}}}
    \\
    &\min\Big(
        \rho_{t,m}^{(i)}(\theta)\hat{A}_m^{(i)},
        \\
    &\quad
        \operatorname{clip}\!\big(
            \rho_{t,m}^{(i)}(\theta),1-\epsilon,1+\epsilon
        \big)\hat{A}_m^{(i)}
    \Big)
    \Bigg],
\end{aligned}
\label{eq:stage_grpo_objective}
\end{equation}
where the expectation is over reset states
$s_0 \sim \rho_m$ and the corresponding rollout groups
generated by the old policy in the world model.
The advantage $\hat{A}_m^{(i)}$ is obtained by normalizing
the local return $R_m^{(i)}=\sum_t r_{t,m}^{(i)}=y_m^{(i)}$
within each group sharing the same stage and reset state;
$\rho_{t,m}^{(i)}(\theta)$ denotes the same surrogate policy
ratio used above.

Together, stage-balanced resets provide direct training
opportunities across all stages, including the initial stage,
while eliminating cross-stage model-generated prefixes.
Stage-specific rewards further allow each skill segment
to be optimized without requiring successful execution of
the remaining task.
All stages optimize a shared VLA policy, which executes
the complete task end-to-end without resets at deployment.

\section{Experiments}

Our experiments evaluate the proposed world model and its use as an
interactive environment for VLA-RL.
We organize the evaluation around five questions:

\begin{enumerate}

    \item \textbf{World Model Capability (Sec.\ref{sec:WM capability})}
    Does the proposed ACWM provide the basic predictive capability required
    of a VLA interaction environment, including high fidelity,
    cross-view consistent, and action-conditioned video generation?

    \item \textbf{Policy-Level Utility of the ACWM (Sec.\ref{sec:WM policy utility})}
    Does the WM, beyond high-fidelity video prediction, support
    reliable policy evaluation and provide useful training data
    for policy improvement?

    \item \textbf{Exploration in WM-RL (Sec.\ref{sec:Exploration in WM-RL})}
    How should a flow-based VLA explore when interacting with the learned environment?

    \item \textbf{Policy Optimization with WM-RL (Sec.\ref{sec:Policy Optimization with WM-RL})}
    Can RL entirely through imagined VLA--WM interaction improve the policy?

    \item \textbf{Long-Horizon Policy Learning (Sec.\ref{sec:long_horizon_manipulation})}
    How can WM-RL be organized to effectively optimize long-horizon tasks composed of multiple sequential manipulation skills?

\end{enumerate}

Our experiments span both simulation and real robots: RoboTwin is used for controlled world-model and WM-RL studies, while policy-level consistency, multi-task WM-RL, and long-horizon task are evaluated on real robots.

\subsection{World Model Capability}
\label{sec:WM capability}
\subsubsection{Video Prediction Setup}
We evaluate video generation on RoboTwin, training on 2,000 trajectories from 50 tasks (40 per task) and testing on 50 held-out trajectories (one per task). We compare against the following representative ACWMs:

\begin{itemize}
    \item \textbf{WoVR~\cite{jiang2026wovrworldmodelsreliable}.}
    WoVR serves as the methodological basis substantially extended in this work. 
    We evaluate both the original single-view model and a three-view adaptation.
    For the three-view adaptation, WoVR is trained on the same three-view observations and absolute-joint commands as our model.
    \item \textbf{Ctrl-World~\cite{CtrlWorld}.}
    Ctrl-World is an SVD-based~\cite{blattmann2023stablevideodiffusionscaling} ACWM that natively supports
    multi-view video prediction and conditions generation on robot actions through cross-attention.
    
    \item \textbf{EVAC~\cite{jiang2025enerverseacenvisioningembodiedenvironments}.}
    EVAC adopts a U-Net~\cite{ronneberger2015unetconvolutionalnetworksbiomedical} video-generation architecture and explicitly
    represents robot and camera geometry using action maps and camera raymaps, while supporting multi-view prediction.

\end{itemize}

We report LPIPS~\cite{lpips}, FID~\cite{fid}, FVD~\cite{fvd}, SSIM~\cite{SSIM}, and PSNR~\cite{PSNR} as video-quality metrics.

\subsubsection{Multi-View Prediction Fidelity and Efficiency}

\begin{table*}[htbp]
    \centering
    \caption{
        Video prediction quality.
        Wrist metrics are averaged over the two wrist cameras.
        Best and second-best results are shown in bold and underlined,
        respectively. ``--'' denotes an unavailable view.
    }
    \label{tab:wm_video_quality}

    \setlength{\tabcolsep}{3.0pt}
    \renewcommand{\arraystretch}{1.08}

    \begin{tabular}{@{}lc*{5}{c}*{5}{c}c@{}}
        \toprule
        \multirow{2}{*}{\textbf{Method}}
        & \multirow{2}{*}{\textbf{NFE}}
        & \multicolumn{5}{c}{\textbf{Head View}}
        & \multicolumn{5}{c}{\textbf{Wrist Views}}
        & \multirow{2}{*}{\textbf{FPS}$\uparrow$}
        \\
        \cmidrule(lr){3-7}
        \cmidrule(lr){8-12}
        &
        & LPIPS$\downarrow$
        & FID$\downarrow$
        & FVD$\downarrow$
        & SSIM$\uparrow$
        & PSNR$\uparrow$
        & LPIPS$\downarrow$
        & FID$\downarrow$
        & FVD$\downarrow$
        & SSIM$\uparrow$
        & PSNR$\uparrow$
        &
        \\
        \midrule

        Ctrl-World
        & 5
        & 0.318
        & 73.22
        & 229.83
        & 0.812
        & 16.39
        & 0.350
        & 117.45
        & 305.39
        & 0.863
        & 22.38
        & 4.16
        \\

        & 50
        & 0.305
        & 47.16
        & 147.46
        & 0.802
        & 15.86
        & 0.346
        & 73.67
        & 186.66
        & 0.849
        & 22.14
        & 0.53
        \\

        \addlinespace[1pt]

        EVAC
        & 27
        & 0.337
        & 50.93
        & 196.72
        & 0.768
        & 13.76
        & 0.328
        & 68.49
        & 185.15
        & 0.833
        & 20.89
        & 1.71
        \\

        \midrule

        WoVR (1 view)
        & 5
        & 0.182
        & \underline{20.74}
        & 75.09
        & 0.858
        & 19.12
        & --
        & --
        & --
        & --
        & --
        & 22.56
        \\

        WoVR (3 views)
        & 5
        & 0.231
        & 27.01
        & 109.33
        & 0.825
        & 16.96
        & 0.269
        & 47.55
        & 124.41
        & 0.825
        & 23.16
        & 14.01
        \\

        \addlinespace[1pt]
        
        \textbf{WoVR+}
        & 5
        & \textbf{0.148}
        & 22.45
        & 62.16
        & \textbf{0.883}
        & \textbf{22.28}
        & \underline{0.204}
        & \underline{44.10}
        & \underline{81.33}
        & \textbf{0.894}
        & \textbf{29.77}
        & 9.41
        \\

        & 50
        & \underline{0.153}
        & \textbf{19.92}
        & \textbf{49.61}
        & \underline{0.877}
        & 21.86
        & \textbf{0.203}
        & \textbf{35.96}
        & \textbf{65.77}
        & 0.888
        & \underline{29.39}
        & 1.78
        \\

        \addlinespace[1pt]

        \textbf{WoVR+(DMD)}
        & 1
        & 0.163
        & 23.59
        & \underline{57.41}
        & \underline{0.877}
        & \underline{22.20}
        & 0.236
        & 53.58
        & 106.15
        & \underline{0.889}
        & 28.79
        & 15.03
        \\

        \bottomrule
    \end{tabular}%
\end{table*}

\begin{figure}[bthp]
\centering
\includegraphics[width=\linewidth]{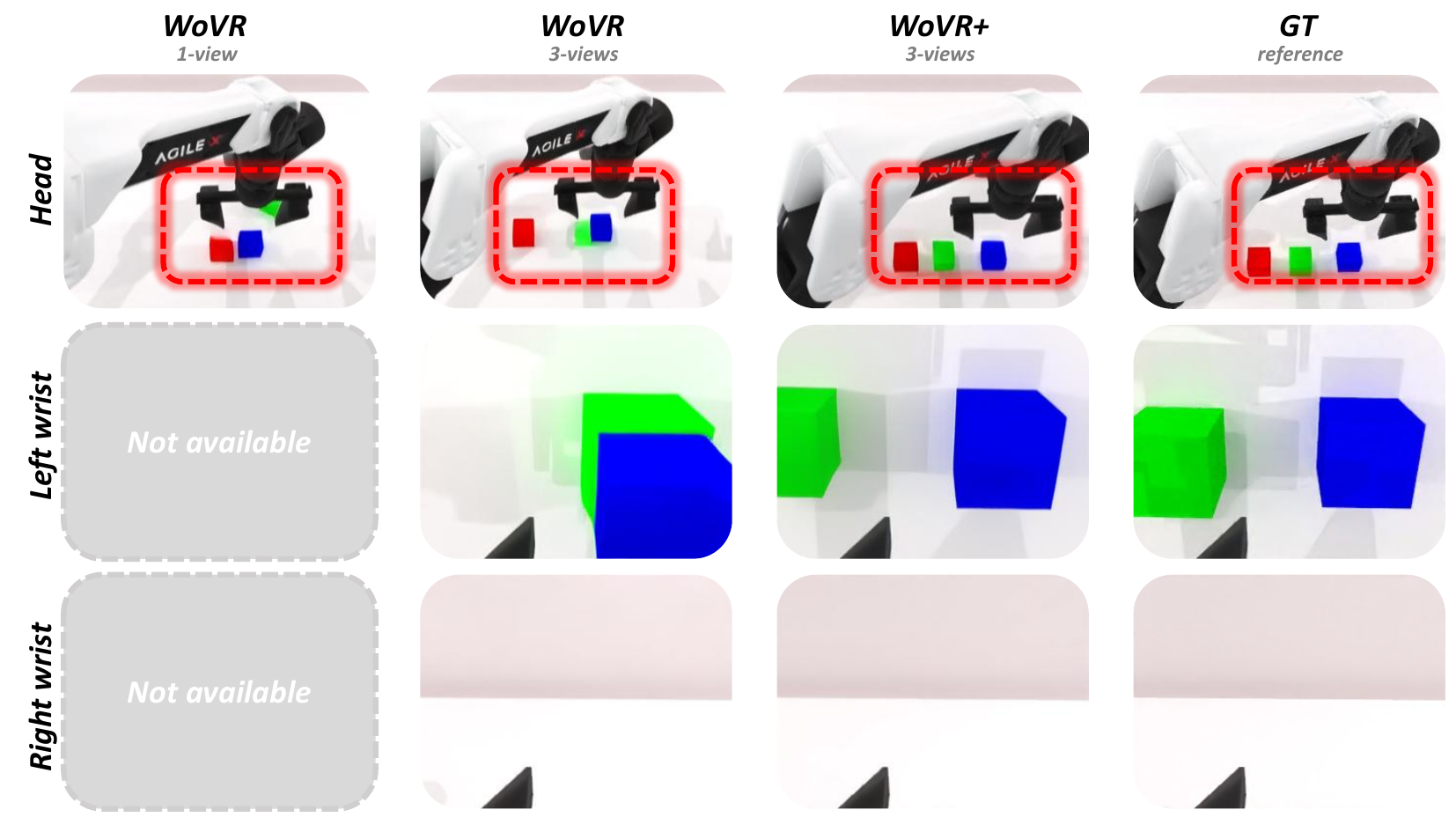}
\caption{\textbf{Qualitative comparison of multi-view prediction.
}}
\label{fig:CV1}
\end{figure}

Table~\ref{tab:wm_video_quality} shows  our 5-step model achieves the best overall fidelity across head and wrist views with high throughput, while DMD2 enables one-step generation. As shown in Fig.~\ref{fig:CV1}, although extending WoVR to three views improves localization, its cross-view layouts remain inconsistent; our model better preserves object counts and spatial relations.

\subsubsection{Autoregressive Prediction Fidelity}
We next evaluate the first 200 frames of 20 held-out autoregressive rollouts in eight 25-frame intervals, reporting frame-aligned LPIPS, SSIM, and PSNR since each interval contains too few samples for reliable FID or FVD.

\begin{figure}[htbp]
    \centering
    \includegraphics[width=\linewidth]{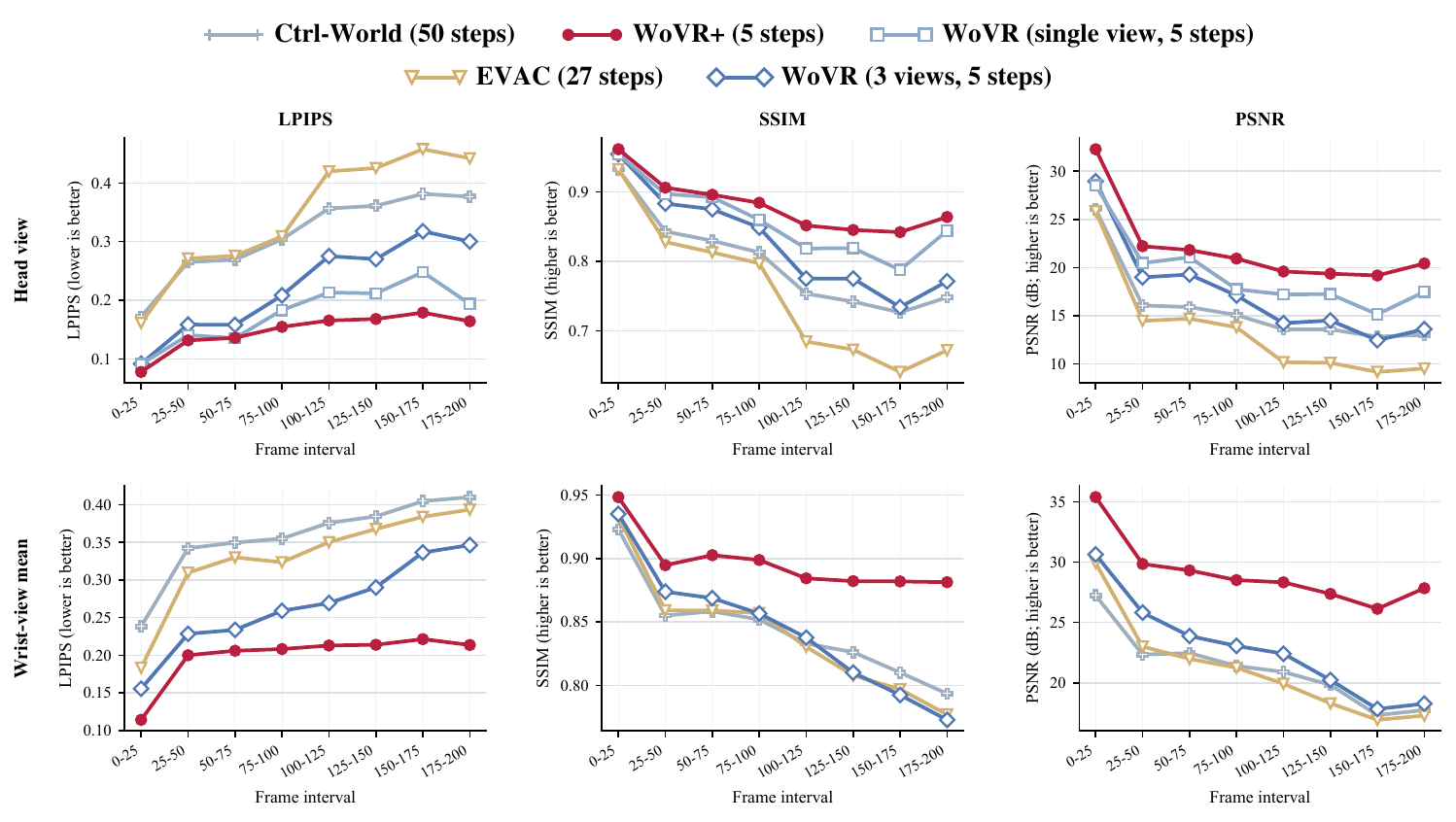}
    \par\vspace{2mm} 
    \includegraphics[width=\linewidth]{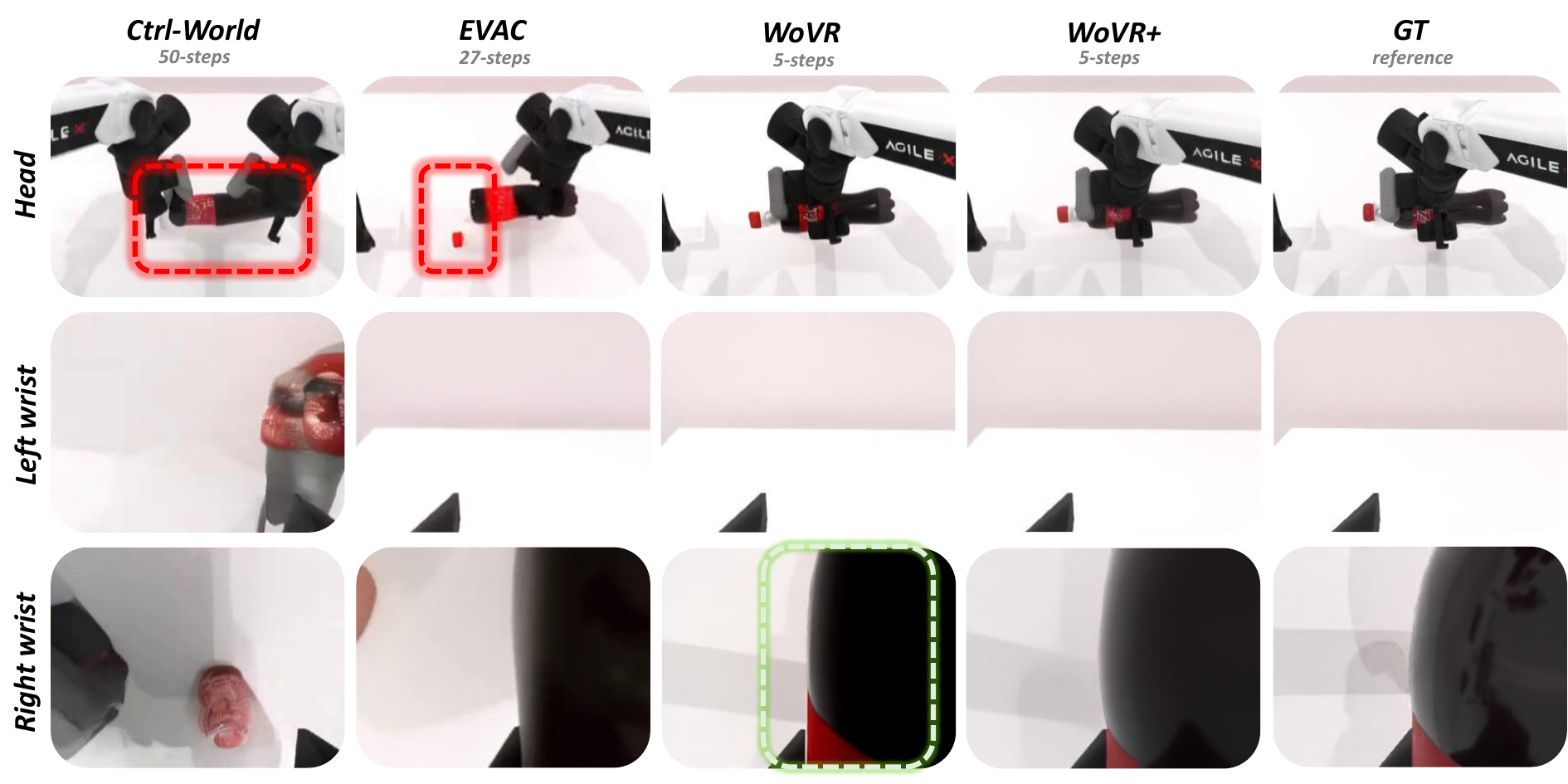}
    \caption{\textbf{Long-horizon prediction fidelity over a fixed cohort of 20 held-out trajectories.}}
    \label{fig:long_horizon_fidelity}
    \vspace{-4mm}
\end{figure}

Fig.~\ref{fig:long_horizon_fidelity} shows that all methods degrade over time, while ours maintains the best overall fidelity across views and horizons. This motivates limiting long model-generated prefixes despite the improved autoregressive stability. Qualitatively, Ctrl-World and EVAC exhibit structural distortions, while WoVR produces less realistic surface appearance; our WoVR+ better preserves both geometry and visual realism.

\subsubsection{Prediction under Policy-Induced Actions}
Held-out demonstration actions do not capture the action shift encountered in WM-RL, where policy exploration produces action chunks that are executable but underrepresented in the training data. We therefore sample Flow-SDE actions from held-out RoboTwin states, execute them to obtain ground-truth futures, and condition each world model on the same history and actions. As shown in Table~\ref{tab:exploratory_action_fidelity}, our model outperforms all three-view baselines, demonstrating stronger local generalization under policy-induced exploration, consistent with its image-space action grounding and explicit camera conditioning.

\begin{table*}[htbp]
    \centering
    \caption{
        Prediction fidelity under policy-induced action distribution shift.
    }
    \label{tab:exploratory_action_fidelity}

    \setlength{\tabcolsep}{3.0pt}
    \renewcommand{\arraystretch}{1.08}

    \begin{tabular}{@{}lc*{5}{c}*{5}{c}@{}}
        \toprule
        \multirow{2}{*}{\textbf{Method}}
        & \multirow{2}{*}{\textbf{NFE}}
        & \multicolumn{5}{c}{\textbf{Head View}}
        & \multicolumn{5}{c}{\textbf{Wrist Views}}
        \\
        \cmidrule(lr){3-7}
        \cmidrule(lr){8-12}
        &
        & LPIPS$\downarrow$
        & FID$\downarrow$
        & FVD$\downarrow$
        & SSIM$\uparrow$
        & PSNR$\uparrow$
        & LPIPS$\downarrow$
        & FID$\downarrow$
        & FVD$\downarrow$
        & SSIM$\uparrow$
        & PSNR$\uparrow$
        \\
        \midrule

        Ctrl-World
        & 5
        & 0.264
        & 80.42
        & 198.17
        & 0.847
        & 15.99
        & 0.311
        & 97.43
        & 491.56
        & \textbf{0.913}
        & 19.57
        \\

        & 50
        & 0.278
        & 51.86
        & 191.81
        & 0.823
        & 14.89
        & 0.338
        & 58.02
        & \underline{229.34}
        & 0.890
        & 18.71
        \\

        \addlinespace[1pt]

        EVAC
        & 27
        & 0.304
        & 56.74
        & 147.66
        & 0.809
        & 14.12
        & \underline{0.291}
        & 56.59
        & 278.99
        & 0.887
        & \underline{20.00}
        \\

        \midrule

        WoVR (1 view)
        & 5
        & \underline{0.228}
        & \textbf{34.58}
        & \underline{131.12}
        & \underline{0.854}
        & \underline{16.36}
        & --
        & --
        & --
        & --
        & --
        \\

        WoVR (3 views)
        & 5
        & 0.236
        & 36.91
        & 144.73
        & 0.847
        & 16.09
        & 0.322
        & \underline{49.34}
        & 297.26
        & 0.856
        & 19.21
        \\

        \addlinespace[1pt]
        
        \textbf{WoVR+}
        & 5
        & \textbf{0.206}
        & \underline{36.18}
        & \textbf{107.22}
        & \textbf{0.878}
        & \textbf{18.55}
        & \textbf{0.265}
        & \textbf{40.23}
        & \textbf{225.84}
        & \underline{0.890}
        & \textbf{22.50}
        \\

        \bottomrule
    \end{tabular}%
    
\end{table*}

\subsubsection{Ablation Study of Model Structure and Training methods}
Having established the overall prediction fidelity and robustness of our world model, we next examine which architectural and training components contribute to these improvements.
We first ablate the geometric-conditioning design and then analyze the training strategies introduced for autoregressive robustness.

For the architectural ablations, we keep the action maps and all
remaining components fixed, and vary only the camera-geometry
conditioning.
We compare removing raymaps entirely, encoding ray origins and
directions as rasterized images followed by VAE encoding
(\emph{Image + VAE}), and constructing them directly on the
video-latent grid (\emph{Direct Latent}), each with or without
learned view embeddings.

As shown in Table~\ref{tab:geometry_ablation}, removing raymaps
produces the weakest performance, while direct latent ray encoding
consistently outperforms the image-based representation.
Adding view embeddings further improves prediction quality,
indicating that continuous camera geometry and explicit view identity provide complementary information beyond EEF projection alone.

\begin{table*}[htbp]
    \centering
    \caption{
        Ablation of camera raymap encoding and view embeddings. 
    }
    \label{tab:geometry_ablation}

    \setlength{\tabcolsep}{3.2pt}
    \renewcommand{\arraystretch}{1.05}

    \begin{tabular}{@{}llc*{5}{c}*{5}{c}@{}}
        \toprule
        \multirow{2}{*}{\textbf{RayMap Encoding}}
        & \multirow{2}{*}{\textbf{View Emb.}}
        & \multirow{2}{*}{\textbf{NFE}}
        & \multicolumn{5}{c}{\textbf{Head View}}
        & \multicolumn{5}{c}{\textbf{Wrist Views}}
        \\
        \cmidrule(lr){4-8}
        \cmidrule(lr){9-13}
        &
        &
        & LPIPS$\downarrow$
        & FID$\downarrow$
        & FVD$\downarrow$
        & SSIM$\uparrow$
        & PSNR$\uparrow$
        & LPIPS$\downarrow$
        & FID$\downarrow$
        & FVD$\downarrow$
        & SSIM$\uparrow$
        & PSNR$\uparrow$
        \\
        \midrule
        \multirow{2}{*}{No RayMap}
        & \multirow{2}{*}{No}
        & 5
        & 0.2335
        & 32.67
        & 101.07
        & 0.8275
        & 16.91
        & 0.2842
        & 61.14
        & 130.27
        & 0.8685
        & 26.57
        \\

        &
        & 50
        & 0.2360
        & 27.46
        & 81.22
        & 0.8243
        & 16.80
        & 0.2875
        & 52.06
        & 114.71
        & 0.8589
        & 26.24
        \\

        \midrule
        
        \multirow{2}{*}{Direct Latent}
        & \multirow{2}{*}{No}
        & 5
        & 0.1829
        & 27.73
        & 79.99
        & 0.8593
        & 20.10
        & 0.2439
        & 53.30
        & 105.46
        & 0.8780
        & 27.81
        \\

        &
        & 50
        & 0.1862
        & 24.36
        & 64.72
        & \textbf{0.8553}
        & 19.82
        & 0.2447
        & \textbf{44.30}
        & 94.14
        & 0.8698
        & 27.71
        \\

        \midrule

        \multirow{2}{*}{Image + VAE}
        & \multirow{2}{*}{No}
        & 5
        & 0.2148
        & 33.29
        & 103.37
        & 0.8433
        & 18.43
        & 0.2634
        & 57.55
        & 120.34
        & 0.8735
        & 27.02
        \\

        &
        & 50
        & 0.2203
        & 27.90
        & 81.76
        & 0.8363
        & 18.03
        & 0.2615
        & 47.70
        & 101.28
        & 0.8664
        & 26.80
        \\

        \midrule

        \multirow{2}{*}{Image + VAE}
        & \multirow{2}{*}{Yes}
        & 5
        & 0.2070
        & 30.74
        & 99.43
        & 0.8456
        & 18.73
        & 0.2552
        & 56.38
        & 115.03
        & 0.8779
        & 26.71
        \\

        &
        & 50
        & 0.2161
        & 26.18
        & 77.92
        & 0.8368
        & 18.25
        & 0.2582
        & 46.23
        & 88.75
        & 0.8636
        & 26.74
        \\

        \midrule

        \multirow{2}{*}{\textbf{Direct Latent}}
        & \multirow{2}{*}{\textbf{Yes}}
        & 5
        & \textbf{0.1801}
        & \textbf{26.79}
        & \textbf{76.29}
        & \textbf{0.8613}
        & \textbf{20.29}
        & \textbf{0.2333}
        & \textbf{53.04}
        & \textbf{101.11}
        & \textbf{0.8875}
        & \textbf{28.29}
        \\

        &
        & 50
        & \textbf{0.1850}
        & \textbf{22.88}
        & \textbf{63.50}
        & 0.8549
        & \textbf{19.96}
        & \textbf{0.2330}
        & 44.49
        & \textbf{87.95}
        & \textbf{0.8783}
        & \textbf{28.13}
        \\

        \bottomrule
    \end{tabular}%
\end{table*}

We next fix the full geometric-conditioning architecture and ablate the training strategies introduced for robust autoregressive prediction.
As shown in Table~\ref{tab:wm_training_ablation}, all three strategies improve generation quality over the raw training configuration.
Anchored sparse history provides a substantial gain by preserving both the initial scene context and recent task progress, while geometry-aware local corruption improves robustness to imperfect generated context. Temporal-stride sampling yields the largest individual improvement, suggesting that exposing the model to different motion magnitudes and transition intervals is particularly important for action-conditioned prediction.
Combining all three strategies achieves the strongest distribution-level and wrist-view performance, although temporal-stride sampling alone remains slightly better on several frame-aligned head-view metrics.

\begin{table*}[htbp]
    \centering
    \caption{
        Ablation of world-model training strategies. Each individual variant adds the indicated strategy to the raw configuration, while \emph{All} combines all the strategies. 
    }
    \label{tab:wm_training_ablation}

    \setlength{\tabcolsep}{3.2pt}
    \renewcommand{\arraystretch}{1.05}

    \begin{tabular}{@{}lc*{5}{c}*{5}{c}@{}}
        \toprule
        \multirow{2}{*}{\textbf{Training Configuration}}
        & \multirow{2}{*}{\textbf{NFE}}
        & \multicolumn{5}{c}{\textbf{Head View}}
        & \multicolumn{5}{c}{\textbf{Wrist Views}}
        \\
        \cmidrule(lr){3-7}
        \cmidrule(lr){8-12}
        &
        & LPIPS$\downarrow$
        & FID$\downarrow$
        & FVD$\downarrow$
        & SSIM$\uparrow$
        & PSNR$\uparrow$
        & LPIPS$\downarrow$
        & FID$\downarrow$
        & FVD$\downarrow$
        & SSIM$\uparrow$
        & PSNR$\uparrow$
        \\
        \midrule

        \multirow{2}{*}{Raw}
        & 5
        & 0.1801
        & 26.79
        & 76.29
        & 0.8613
        & 20.29
        & 0.2333
        & 53.04
        & 101.11
        & 0.8875
        & 28.29
        \\

        & 50
        & 0.1850
        & 22.88
        & 63.50
        & 0.8549
        & 19.96
        & 0.2330
        & 44.49
        & 87.95
        & 0.8783
        & 28.13
        \\

        \midrule

        \multirow{2}{*}{+ Sparse History}
        & 5
        & 0.1623
        & 25.52
        & 73.77
        & 0.8743
        & 21.33
        & 0.2263
        & 48.57
        & 95.57
        & 0.8898
        & 29.05
        \\

        & 50
        & 0.1660
        & 21.03
        & 55.35
        & 0.8689
        & 21.04
        & 0.2254
        & 39.71
        & 77.52
        & 0.8817
        & 28.82
        \\

        \midrule

        \multirow{2}{*}{+ Geometry-Aware Blur}
        & 5
        & 0.1721
        & 25.82
        & 74.94
        & 0.8681
        & 20.92
        & 0.2288
        & 49.21
        & 92.84
        & 0.8885
        & 28.69
        \\

        & 50
        & 0.1767
        & 21.82
        & 57.01
        & 0.8627
        & 20.57
        & 0.2308
        & 40.34
        & 82.28
        & 0.8803
        & 28.52
        \\

        \midrule

        \multirow{2}{*}{+ Temporal Stride}
        & 5
        & \textbf{0.1473}
        & 23.11
        & 66.51
        & \textbf{0.8859}
        & \textbf{22.50}
        & 0.2058
        & 45.74
        & 82.16
        & \textbf{0.8964}
        & \textbf{29.86}
        \\

        & 50
        & \textbf{0.1529}
        & 20.01
        & 51.28
        & \textbf{0.8789}
        & \textbf{21.95}
        & 0.2100
        & 38.47
        & 70.42
        & 0.8867
        & 29.34
        \\

        \midrule

        \multirow{2}{*}{\textbf{All}}
        & 5
        & 0.1478
        & \textbf{22.45}
        & \textbf{62.16}
        & 0.8830
        & 22.28
        & \textbf{0.2041}
        & \textbf{44.10}
        & \textbf{81.33}
        & 0.8944
        & 29.77
        \\

        & 50
        & 0.1533
        & \textbf{19.92}
        & \textbf{49.61}
        & 0.8771
        & 21.86
        & \textbf{0.2029}
        & \textbf{35.96}
        & \textbf{65.77}
        & \textbf{0.8878}
        & \textbf{29.39}
        \\

        \bottomrule
    \end{tabular}%
\end{table*}

\subsection{Policy-Level Utility of the Learned Environment}
\label{sec:WM policy utility}
The preceding section evaluates predictive fidelity at the observation level. We next ask whether these predictions preserve useful policy-level information before applying RL.

\subsubsection{Policy-Level Consistency}
\begin{figure}[htbp]
    \centering
    \includegraphics[width=\linewidth]{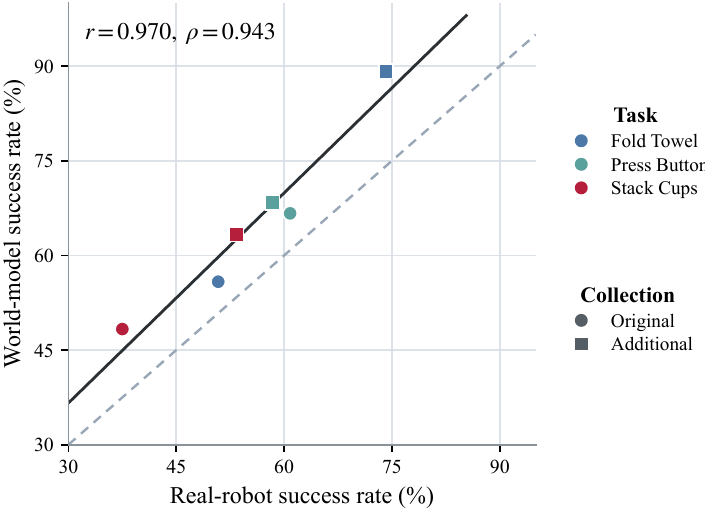}
    \caption{\textbf{Real-robot and world-model success rates across six
    task--collection groups.} Colors denote tasks; circles and squares
    distinguish different checkpoints.}
    \label{fig:policy_consistency_six}
    \vspace{-4mm}
\end{figure}

We evaluate two VLA policies on each of three tasks, conducting
120 rollouts in the ACWM and 120 on the real robot for each
task--policy pair. Each ACWM rollout is initialized using the
initial images from the corresponding real-robot rollout.
As shown in Fig~\ref{fig:policy_consistency_six}, success rates in the learned and real environments are strongly correlated across the six task–policy pairs
(Pearson \(r=0.97\), Spearman \(\rho=0.94\)).

\subsubsection{Synthetic-Data Utility}

We further evaluate whether imagined trajectories provide useful supervision for policy improvement.
Using 500 environment rollouts to train the ACWM, we generate 512 synthetic trajectories from matched reset states.
After success filtering, 401 real and 480 synthetic trajectories are retained, forming an 881-trajectory dataset for continued SFT.
WM-BC improves success from 71.00\% to 91.00\% under fixed resets and from 72.66\% to 76.56\% under randomized resets.
The smaller gain under randomized initialization suggests that filtered synthetic demonstrations provide useful but incomplete supervision, motivating direct WM-RL over both successful and failed rollouts.

\subsection{Exploration in WM-RL}
\label{sec:Exploration in WM-RL}
We next study how the temporal structure of Flow-SDE exploration affects policy optimization in the learned environment. We first compare I.I.D. and chunk-wise exploration under otherwise matched WM-RL configurations, and then analyze the resulting action perturbations and executed task-space trajectories to understand the observed optimization behavior.

\subsubsection{Effect on WM-RL Optimization}

\begin{table}[!t]
    \centering
    \caption{
        Success rates (\%) using I.I.D.\ and chunk-wise Flow-SDE noise.
        Fixed evaluation uses 100 shared initial states. 
    }
    \label{tab:noise_ablation}

    \setlength{\tabcolsep}{4.0pt}
    \renewcommand{\arraystretch}{1.06}

    \begin{tabular}{@{}lccccc@{}}
        \toprule
        \multirow{2}{*}{\textbf{Exploration}}
        & \multirow{2}{*}{\textbf{Updates}}
        & \multicolumn{2}{c}{\textbf{Adjust Bottle}}
        & \multicolumn{2}{c}{\textbf{Click Bell}}
        \\
        \cmidrule(lr){3-4}
        \cmidrule(lr){5-6}
        &
        & Fixed
        & Random
        & Fixed
        & Random
        \\
        \midrule

        Initial SFT
        & 0
        & 79.00
        & 70.70
        & 71.00
        & 72.66
        \\

        \midrule

        \multirow{5}{*}{Chunk-wise}
        & 10
        & 76.00
        & 76.95
        & 87.00
        & 81.25
        \\

        & 20
        & 80.00
        & 80.86
        & 89.00
        & 86.72
        \\

        & 30
        & 85.00
        & 86.33
        & 87.00
        & 86.33
        \\

        & 40
        & 86.00
        & 88.28
        & 88.00
        & 89.45
        \\

        & 50
        & \textbf{93.00}
        & \textbf{91.80}
        & \textbf{92.00}
        & \textbf{93.75}
        \\

        \midrule

        \multirow{5}{*}{I.I.D.}
        & 10
        & 72.00
        & 65.63
        & 78.00
        & 75.78
        \\

        & 20
        & 73.00
        & 77.34
        & 71.00
        & 68.36
        \\

        & 30
        & 77.00
        & 78.52
        & 72.00
        & 74.22
        \\

        & 40
        & 70.00
        & 68.36
        & 81.00
        & 74.61
        \\
        
        & 50
        & 77.00
        & 74.22
        & 89.00
        & 86.33
        \\

        \bottomrule
    \end{tabular}%
\end{table}

We first isolate the effect of exploration noise by fixing the initial policy, world model, reward model, rollout budget, and GRPO
configuration, varying only the temporal correlation of the Flow-SDE
innovation.

As shown in Table~\ref{tab:noise_ablation}, chunk-wise exploration
produces more stable policy improvement and consistently achieves higher final success rates than I.I.D. exploration across both tasks and evaluation settings.

\subsubsection{Temporal and Task-Space Analysis}
To explain the optimization advantage observed above, we further analyze how the two Flow-SDE variants differ in both action perturbations and executed EEF motion.

We characterize exploration using EEF ensemble spread, coherent action exploration, and within-chunk jitter.
For replayed episode \(i\), let 
\(\mathbf q_{i,t}=\mathbf p_{i,t}-\mathbf p_{i,0}\) denote the EEF trajectory aligned to its initial position.
Grouping episodes by the active arm \(a\in\{L,R\}\), we define
\begingroup
\allowdisplaybreaks[1]
\begin{align}
S_{\mathrm{EEF}}
&=
\frac{1}{H}\sum_{t=1}^{H}
\sum_{a\in\{L,R\}}\frac{N_a}{N}
\sqrt{
\frac{1}{N_a}\sum_{i\in\mathcal I_a}
\|\mathbf q_{i,t}-\bar{\mathbf q}_{a,t}\|_2^2
}
\notag\\
\Delta\mathbf A_{i,c}
&=
\mathbf A^{\mathrm{SDE}}_{i,c}
-\mathbf A^{\mathrm{ODE}}_{i,c},
\qquad
\boldsymbol{\mu}_{i,c}
=
\frac{1}{T}\sum_{\tau=0}^{T-1}
\Delta\mathbf A_{i,c,\tau}
\notag\\
E_{i,c}
&=
\sqrt{\frac{1}{D}\|\boldsymbol{\mu}_{i,c}\|_2^2}
\notag\\*
J_{i,c}
&=
\sqrt{
\frac{1}{TD}
\sum_{\tau=0}^{T-1}
\|\Delta\mathbf A_{i,c,\tau}
-\boldsymbol{\mu}_{i,c}\|_2^2
}.
\label{eq:exploration_metrics}
\end{align}
\endgroup
Here, $H$ denotes the execution horizon and $T$ is the action-chunk length where \(E\) and \(J\) are averaged over chunks and episodes.
For visualization, the three metrics are min--max normalized over the
evaluated configurations, and the plotted jitter-suppression coordinate is
\(1-\widetilde{J}\).

\begin{figure}[bthp]
\centering
\includegraphics[width=\linewidth]{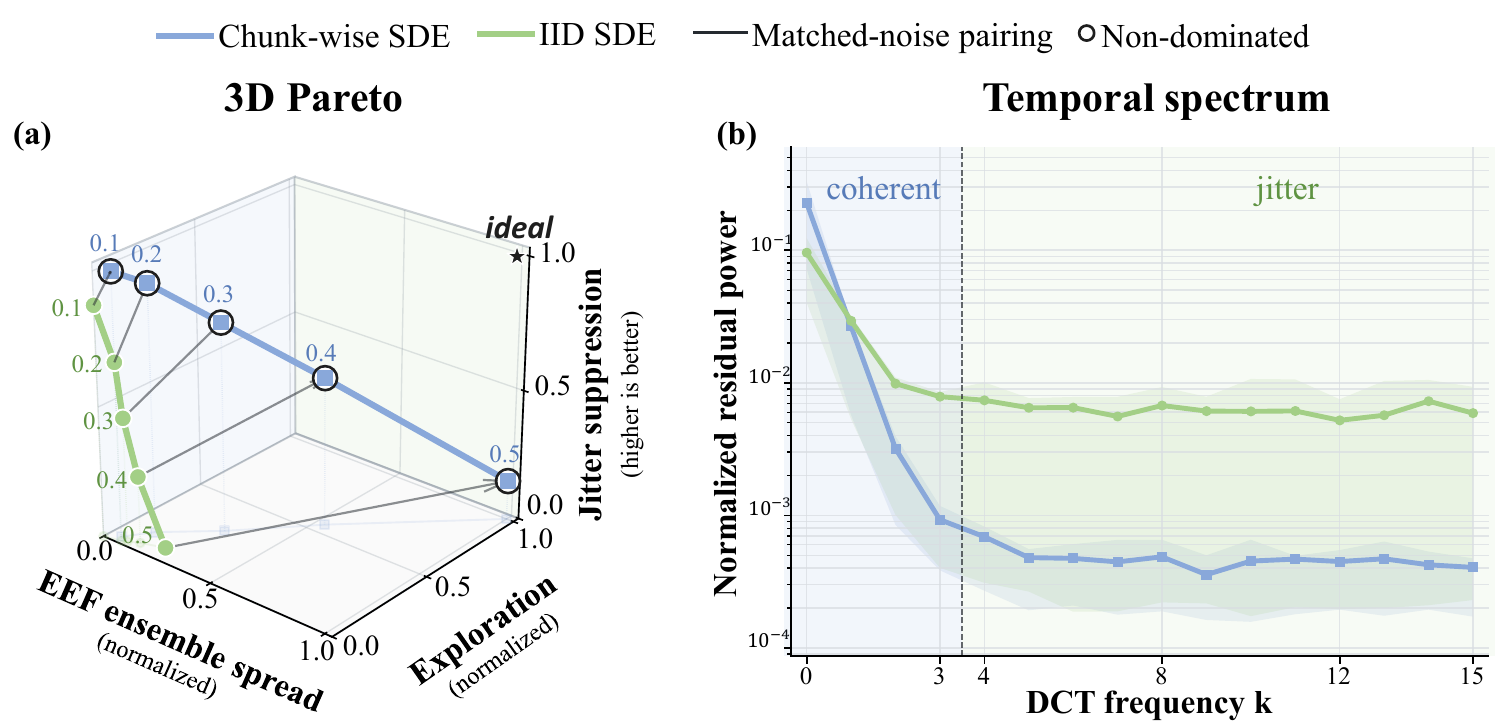}
\caption{
    \textbf{Chunk-wise and I.I.D. Flow-SDE exploration.}
    (a) Across the evaluated noise scales, chunk-wise exploration achieves a better trade-off among EEF ensemble spread, coherent exploration magnitude, and jitter suppression.
    (b) DCT spectra of standardized action residuals show that chunk-wise exploration concentrates perturbations in low-frequency components and exhibits substantially less high-frequency power than I.I.D. exploration.
}
\label{fig:action_exploration_analysis}
\end{figure}

Across the evaluated noise scales, chunk-wise Flow-SDE achieves a better trade-off among EEF ensemble spread, coherent exploration magnitude, and jitter suppression (Fig.~\ref{fig:action_exploration_analysis}).
To further characterize the temporal structure of the perturbations, we apply a DCT along the within-chunk time axis of the standardized action residuals.
I.I.D. noise exhibits substantially more high-frequency power, whereas chunk-wise noise concentrates perturbations in temporally coherent components.
This shows that its advantage arises from reshaping the temporal structure of exploration rather than simply reducing its magnitude.


\begin{figure}[htbp]
    \centering
    \includegraphics[width=\linewidth]{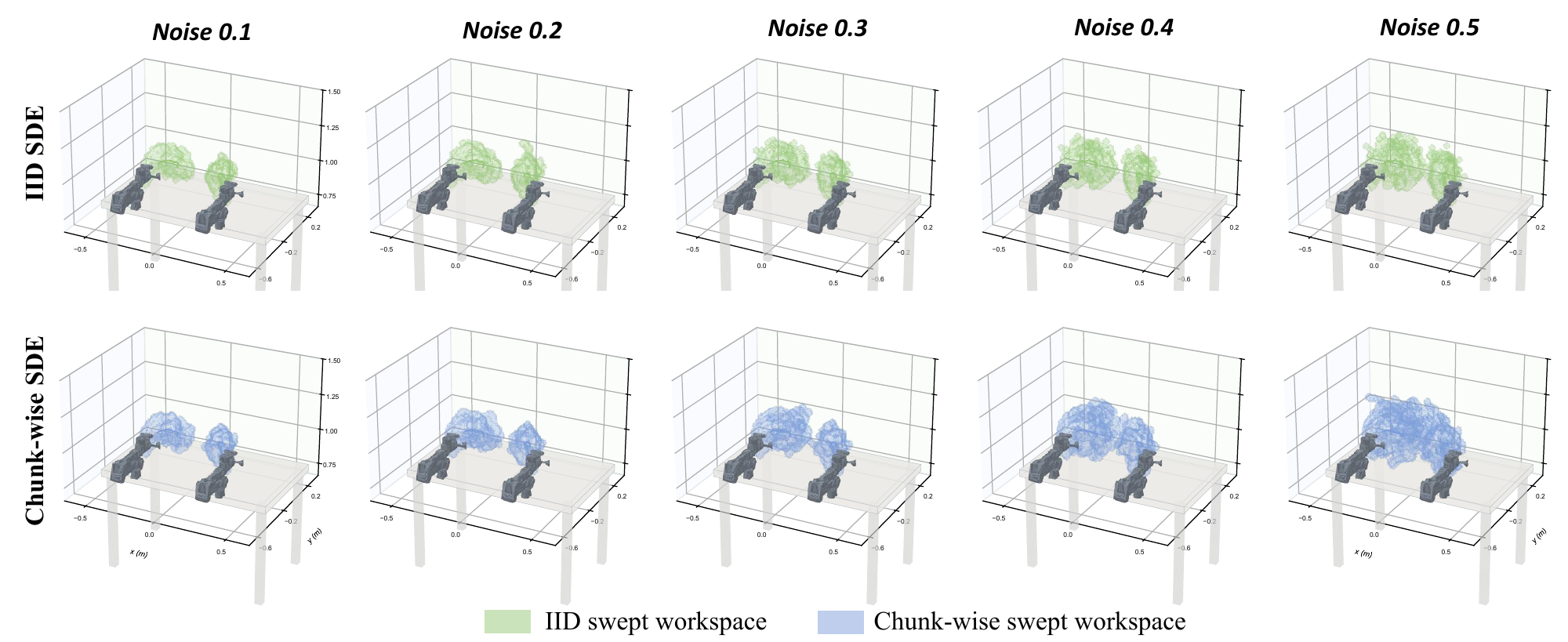}
    \caption{
        \textbf{Executed EEF swept workspace.}
    }
    \label{fig:eef_swept_workspae}
\end{figure}

We further quantify the spatial coverage of the executed EEF trajectories. As shown in Fig.~\ref{fig:eef_swept_workspae}, chunk-wise and I.I.D. Flow-SDE yield similar EEF swept volumes at low noise, while chunk-wise exploration achieves broader coverage as the noise scale increases. This suggests that temporally coherent perturbations convert larger exploration amplitudes into meaningful motion rather than high-frequency oscillation, supporting more stable and effective WM-RL.

\subsection{Policy Optimization with WM-RL}
\label{sec:Policy Optimization with WM-RL}
We evaluated our method on two RoboTwin tasks and six real-robot tasks, with the experiments implemented based on the RLinf framework~\cite{rlinf}.

\subsubsection{RoboTwin Policy Optimization}
Having established chunk-wise Flow-SDE as the exploration strategy,
we next evaluate whether the learned environment itself can support
effective policy optimization.
To isolate the effect of the world model, we fix the SFT policy,
reset states, chunk-wise exploration, reward model, and GRPO
configuration, and vary only the RL environment.
We evaluate both single- and multi-view VLAs; the corresponding base
policies are trained with the same number of demonstrations, and all
optimization settings are matched within each view configuration.

\begin{table}[htbp]
    \centering
    \caption{
        VLA success rates on RoboTwin (\%).
        Fixed uses 100 fixed initial states and Random uses 256 random states.
    }
    \label{tab:single_multi_view_wmrl}

    \footnotesize
    \setlength{\tabcolsep}{4pt}
    \renewcommand{\arraystretch}{1.08}

    \begin{tabular}{@{}lcccc@{}}
        \toprule
        \multirow{2}{*}{\textbf{RL Environment}}
        & \multicolumn{2}{c}{\textbf{Adjust Bottle}}
        & \multicolumn{2}{c}{\textbf{Click Bell}} \\
        \cmidrule(lr){2-3}
        \cmidrule(lr){4-5}
        & Fixed & Random & Fixed & Random \\
        \midrule

        \multicolumn{5}{@{}l}{\textit{Single-view VLA}} \\
        None (SFT)
        & 68.00 & 58.98 & 59.00 & 40.63 \\
        Direct RL (matched)
        & 69.00 & 60.16 & 60.00 & 37.89 \\
        Direct RL (converged)
        & 78.00 & 76.95 & 89.00 & 83.59 \\
        WoVR
        & 69.00 & 63.67 & 77.00 & 68.36 \\
        \textbf{WoVR+ (1 view)}
        & \textbf{75.00}
        & \textbf{71.09}
        & \textbf{85.00}
        & \textbf{73.44} \\

        \midrule

        \multicolumn{5}{@{}l}{\textit{Multi-view VLA}} \\
        None (SFT)
        & 79.00 & 70.70 & 71.00 & 72.66 \\
        Direct RL (matched)
        & 74.00 & 68.75 & 82.00 & 79.30 \\
        Direct RL (converged)
        & 95.00 & 94.53 & 99.00 & 98.83 \\
        Ctrl-World
        & 79.00 & 76.56 & 84.00 & 78.52 \\
        WoVR (3 views)
        & 90.00 & 83.20 & \textbf{92.00} & 90.63 \\
        \textbf{WoVR+}
        & \textbf{93.00}
        & \textbf{91.80}
        & \textbf{92.00}
        & \textbf{93.75} \\
        \bottomrule
    \end{tabular}%
\end{table}

As shown in Table~\ref{tab:single_multi_view_wmrl}, WoVR+ achieves the strongest overall performance among the ACWM-based environments and outperforms budget-matched Direct RL.
It further approaches converged Direct RL, which is trained without
the online-interaction budget constraint.
Together with the preceding exploration ablation, these results show
that chunk-wise exploration supports effective WM-RL, while the
improved learned environment provided by WoVR+ further translates into stronger downstream policy optimization. Moreover, the multi-view policy consistently outperforms its single-view counterpart, demonstrating the benefit of preserving the complete observation interface during both policy learning and imagined optimization.

\subsubsection{Real-Robot WM-RL}

\begin{figure}[htbp]
\centering
\includegraphics[width=\linewidth]{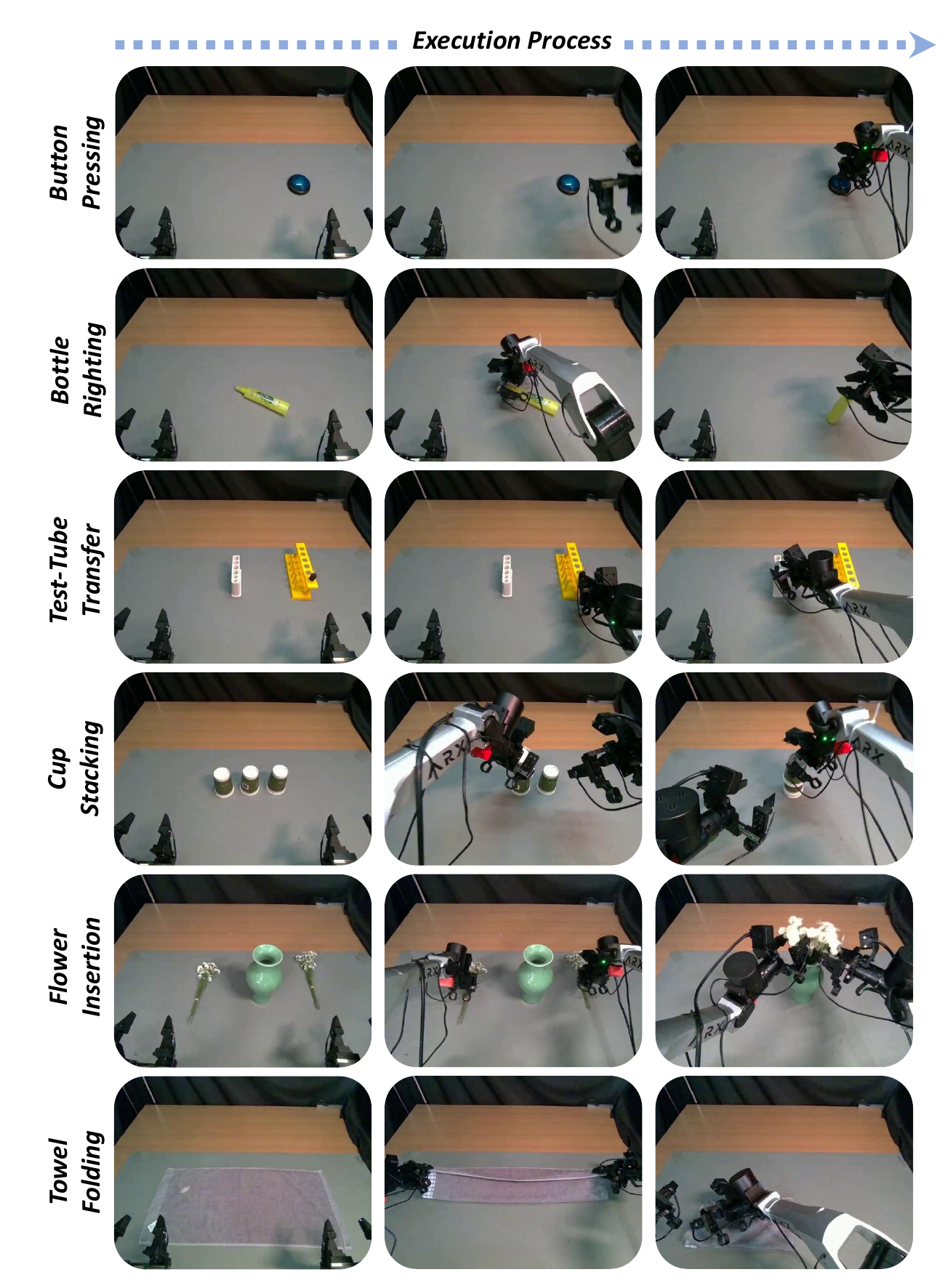}
\caption{\textbf{Real-robot tasks for multi-task WM-RL.} Three single-arm and three bimanual tasks covering contact execution, object
reorientation, precise insertion, multi-object placement, and bimanual deformable-object manipulation}
\label{fig:6tasksettings}
\end{figure}

We next evaluate whether the complete WM-RL pipeline transfers from simulation to physical multi-task manipulation.
We jointly optimize a single language-conditioned $\pi_{0.5}$ policy across three single-arm and three bimanual ARX-5 tasks (Fig.~\ref{fig:6tasksettings}), covering contact execution, object reorientation, precise insertion, multi-object placement, and bimanual deformable-object manipulation.
All tasks share the same observation interface and absolute
joint-and-gripper action space. Details of the real-robot platform, data synchronization, and WM-RL configurations are provided in Appendix~\ref{app:rm}-\ref{app:training_config}.

Starting from SFT with 100 demonstrations per task, we collect 120
real-robot rollouts per task to train a shared world model, task-routed reward models, and data-supported reset pools. Tasks are sampled uniformly during 50 WM-RL updates, with no further real-robot data collection.

\begin{figure*}[htbp]
\centering
\vspace{-4mm}
\includegraphics[width=0.9\textwidth]
{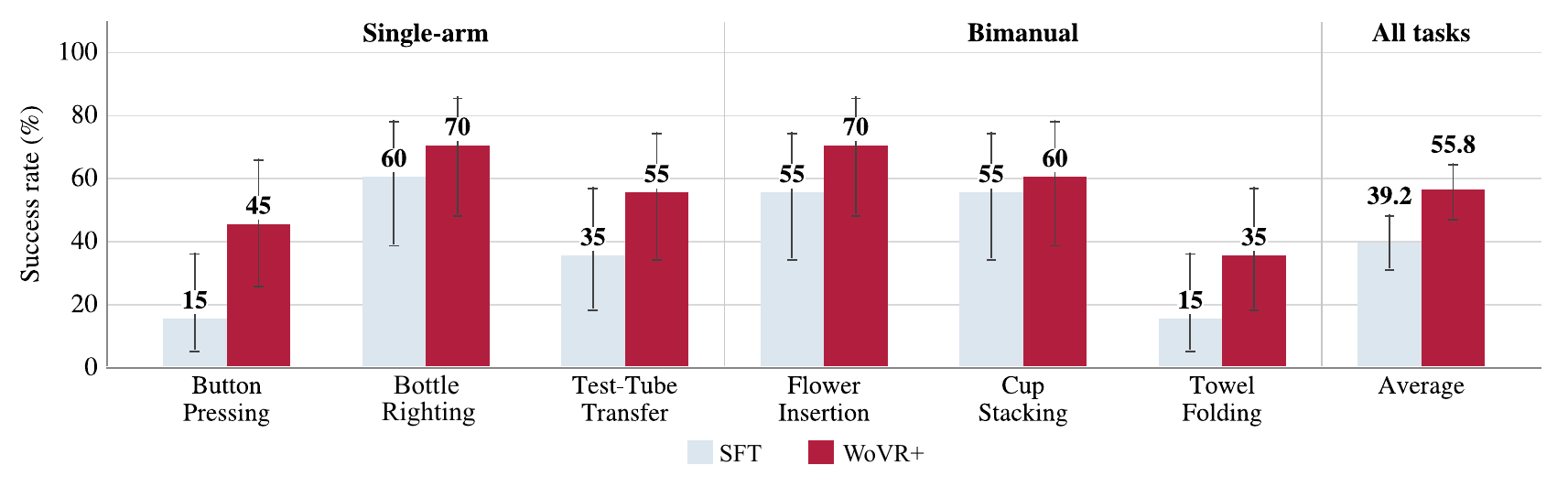}
\caption{\textbf{Real-world multi-task WM-RL.} A single policy is jointly optimized across three single-arm and three bimanual tasks. Bars report success over 20 matched trials per task.
}
\label{fig:real_robot_multitask}
\vspace{-4mm}
\end{figure*}

Across 20 matched real-robot evaluations per task, WM-RL improves
success on all six tasks, increasing the average success rate from
39.2\% to 55.8\% (Fig.~\ref{fig:real_robot_multitask}).
These real-robot results demonstrate that a shared learned environment
can support policy improvements that transfer to physical execution
across heterogeneous single-arm and bimanual tasks within a unified
language-conditioned VLA policy.

\subsection{Long-Horizon Policy Learning}
\label{sec:long_horizon_manipulation}

Long-horizon WM-RL with full-start rollouts is limited by three factors:
\begin{itemize}
    \item \textbf{stage-access imbalance:} later stages receive fewer training opportunities because they depend on successful preceding execution;
    \item \textbf{reward sparsity:} trajectory-level success rewards do not distinguish intermediate task progress;
    \item \textbf{prefix-induced model error:} later-stage optimization relies on increasingly long autoregressive ACWM prefixes.
\end{itemize}

We examine these limitations separately to evaluate whether stage-wise rewards and stage-local imagined training can improve long-horizon policy learning.

\subsubsection{Task, Variants, and Evaluation Protocol}

\begin{figure}[htbp]
    \centering
    \includegraphics[width=0.9\linewidth]
    {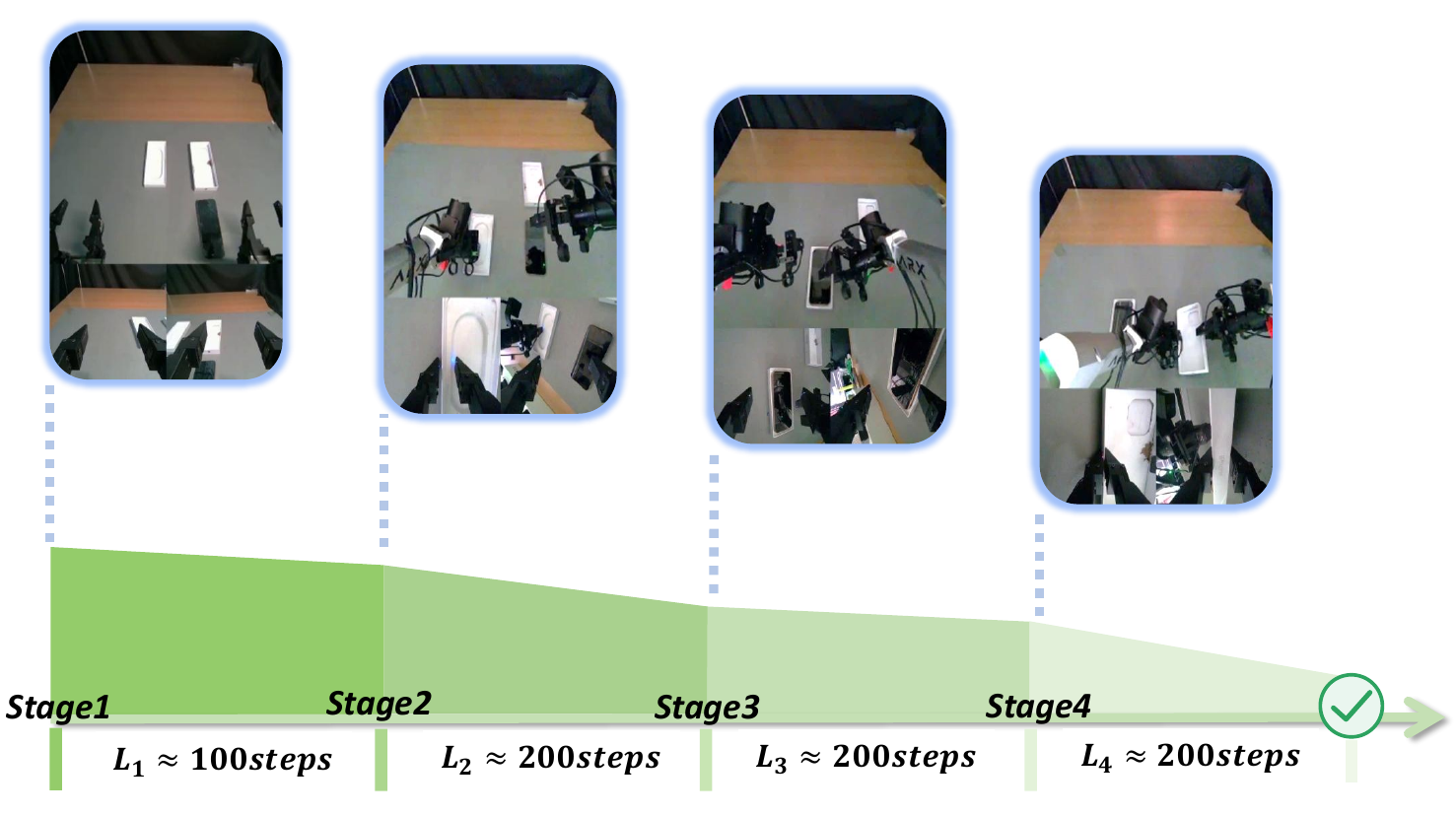}
    \caption{
        \textbf{Stage-reset WM-RL for long-horizon phone packaging.} 4 sequential stages spanning approximately 700 control steps.
        All policies are evaluated from the task initial state
        without intermediate resets.
    }
    \label{fig:phone_packaging}
\end{figure}

We consider \emph{Phone Packaging}, a 4-stage task spanning
approximately 700 control steps at $10\,\mathrm{Hz}$, including base
positioning, phone placement, lid reorientation and handover, and
bimanual lid closure (As shown in Fig.~\ref{fig:phone_packaging}).

Stage boundaries are defined based on transitions between manipulation
skills and observable subgoal completion and finalized through manual
annotation, with one binary reward model for each stage.

Under matched real-data and computing budgets, we evaluate the following WM-RL variants:

\begin{itemize}
    \item \textbf{Full-start + trajectory reward.}
    All rollouts start from the task initial state and receive only a binary full-task success reward.

    \item \textbf{Full-start + stage rewards.}
    Rollouts remain full-start, while stage-specific reward models provide intermediate completion signals along the trajectory.

    \item \textbf{Stage-reset + arrival-matched sampling.}
    Rollouts start from data-supported stage-entry histories, while the stage-sampling probabilities match the empirical stage-arrival distribution of full-start rollouts.

    \item \textbf{Stage-reset + uniform sampling.}
    Stage-entry histories are sampled uniformly across the 4 stages,
    providing balanced local training opportunities.

    \item \textbf{KIR.}
    Following WoVR~\cite{jiang2026wovrworldmodelsreliable} and using a simplified version, imagined rollouts are only initialized from task-level keyframes to shorten the generated prefix.
\end{itemize}

All WM-RL variants optimize the same shared $\pi_{0.5}$ policy under the instruction `Pack the phone.'. 
Stage information is used only during training for reset selection, rollout horizon, and reward routing.
All policies are evaluated end-to-end from the task initial state without intermediate resets.

\subsubsection{Overall Policy Performance}

\begin{figure}[htbp]
    \centering
    \includegraphics[width=\linewidth]{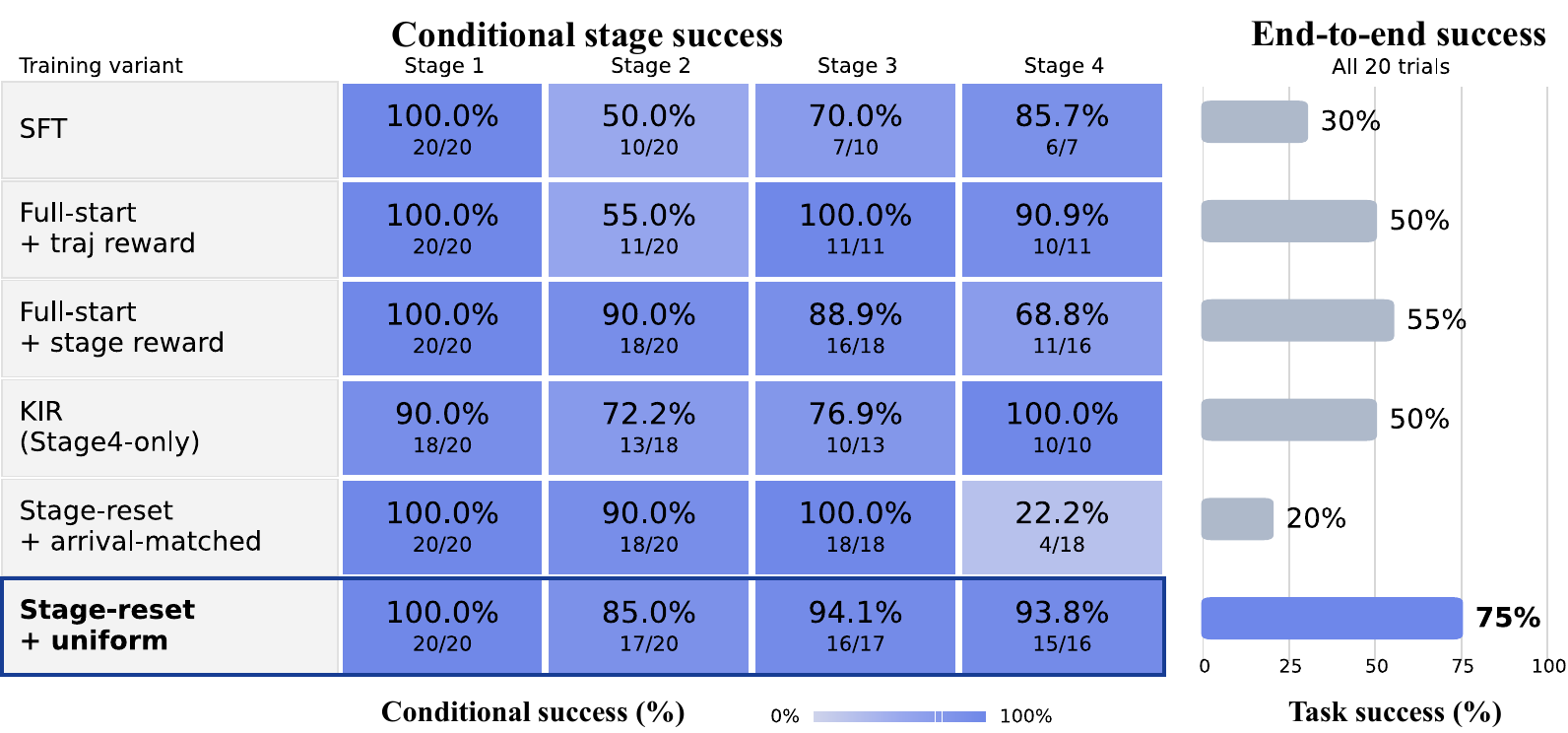}
\caption{
\textbf{Conditional stage success and end-to-end success on the long-horizon Phone Packaging task.}
For each stage, the percentage denotes the conditional probability of completing that stage among trials that successfully reach it; the fraction below reports the corresponding number of successful completions over stage arrivals.
All policies are evaluated from the task initial state without intermediate resets.
}
    \label{fig:stage_conditional}
\end{figure}

Fig~\ref{fig:stage_conditional} reports completion rates for all 4 stages.
SFT succeeds in 30\% of trials, while full-start WM-RL improves this
to 50\% with trajectory-level rewards and 55\% with stage-wise rewards.
KIR reaches 50\%, and stage-reset training shows a strong dependence on the stage-sampling strategy: arrival-matched sampling drops to 20\%, whereas uniform stage sampling achieves the best end-to-end success of 75\%.

The intermediate stage-completion rates further reveal distinct failure patterns.
Full-start stage-wise rewards substantially improve progress through
Stages 2 and 3, but still leave a large gap at Stage 4.
Arrival-matched stage-reset training preserves strong intermediate-stage completion yet fails to convert this into final task success, whereas uniform stage sampling yields the most balanced completion profile across all stages.
We analyze reward granularity, generated-prefix length,
stage-training allocation, and the difference from KIR separately below.

\subsubsection{Effect of Reward Granularity}

We first isolate the effect of reward granularity by keeping full-start imagination unchanged and replacing the trajectory-level binary reward with stage-wise completion rewards.
As shown in Fig.~\ref{fig:stage_conditional}, the end-to-end success improves more moderately from 50\% to 55\%.
This indicates that stage-wise rewards provide substantially more useful supervision for intermediate task progress, but do not by themselves resolve the difficulty of later-stage optimization.

This observation is also reflected in the GRPO training signal:
under the same full-start rollout protocol, approximately 94\% of groups remain informative with stage-wise rewards, compared with 51\% under the trajectory-level binary reward.
Thus, finer-grained rewards alleviate optimization-signal sparsity,
while leaving the state-visitation pattern and world-model prefix length unchanged.

\subsubsection{Effect of Autoregressive Prefix Length}

We next examine how long autoregressive prefixes affect world-model
fidelity at later task stages.
We compare two rollout protocols using the same underlying trajectories:
full-start prediction generates the complete prefix from the task initial state, whereas stage-entry prediction initializes the model directly from recorded observations at the corresponding stage boundary.

\begin{figure}[htbp]
    \centering
    \includegraphics[width=\linewidth]
    {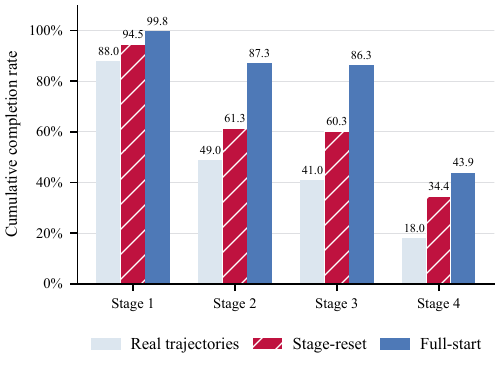}
    \caption{
        \textbf{Stage-completion estimates under full-start and stage-reset imagination.}
        Full-start cumulative completion rates are measured directly, whereas stage-reset cumulative estimates are computed as products of local completion rates.
    }
    \label{fig:stage_calibration}
    \vspace{-8mm}
\end{figure}

Fig.~\ref{fig:stage_calibration} compares the resulting cumulative completion profiles with real trajectories.
Full-start prediction increasingly overestimates later-stage completion, while stage-entry prediction more closely follows the empirical profile.

\begin{figure}[htbp]
    \centering
    \includegraphics[width=\linewidth]
    {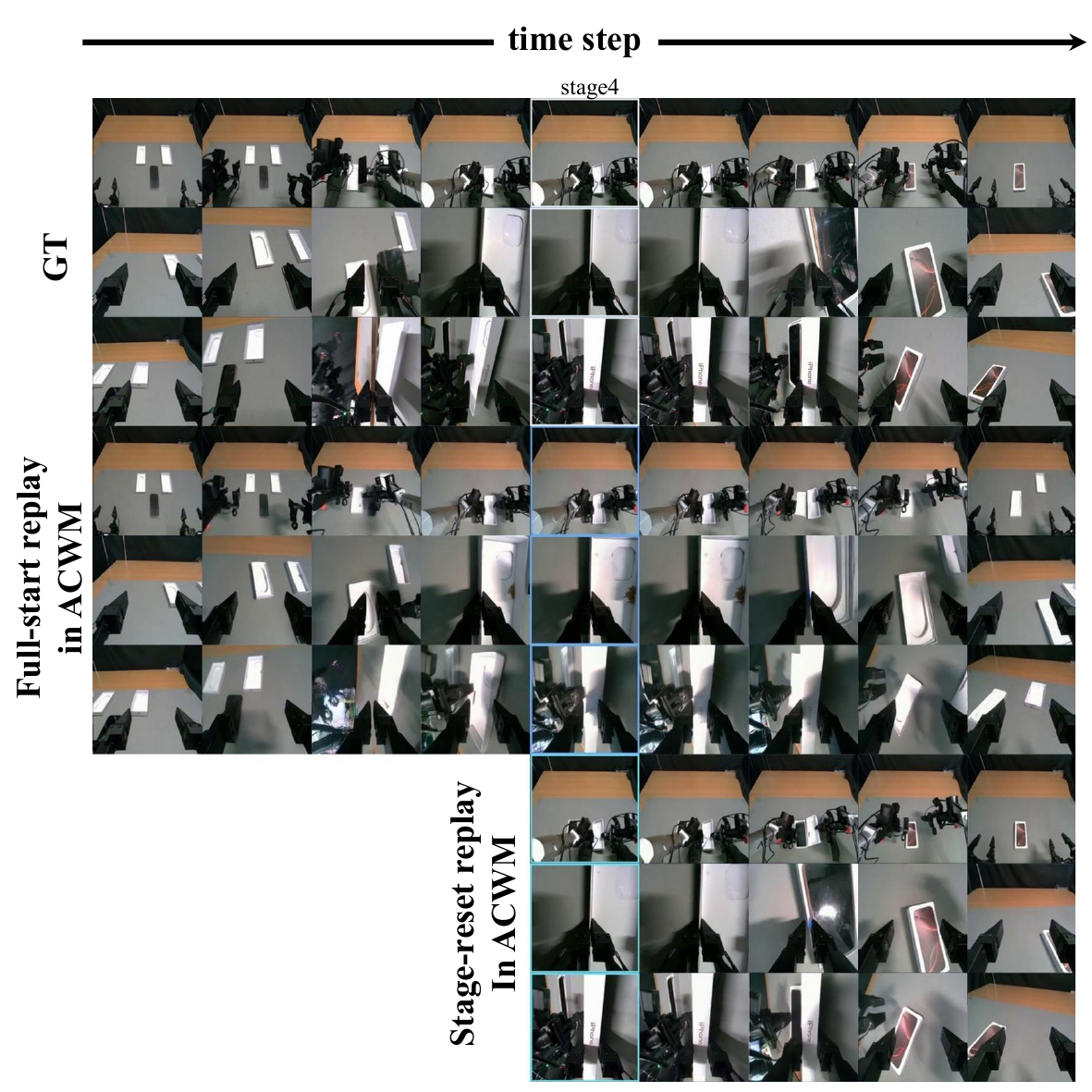}
    \caption{
        \textbf{Effect of generated-prefix length on ACWM replay.}
        Actions from the same recorded trajectory are replayed
        from either the task initial state or the beginning of
        Stage~4.
        Full-start replay exhibits object disappearance and
        positional drift, while the stage-entry reset avoids
        the long generated prefix.
    }
    \label{fig:replay}
\end{figure}

This discrepancy is consistent with the qualitative replay in
Fig.~\ref{fig:replay}, where long-prefix generation exhibits object drift and disappearance, whereas direct stage-entry initialization avoids accumulating errors from the preceding trajectory.
Together, these results show that long autoregressive prefixes distort later-stage world-model predictions, motivating local initialization for long-horizon imagined interaction.

\begin{figure}[htbp]
    \centering
    \includegraphics[width=\linewidth]
{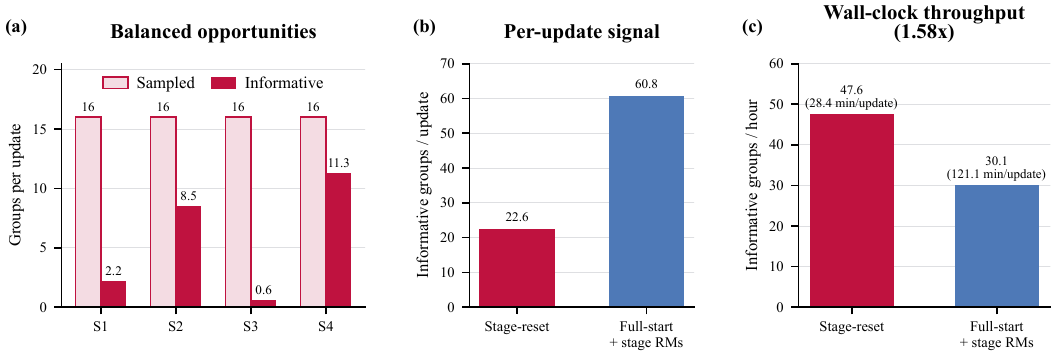}
    \caption{
        \textbf{Informative-group generation under stage-reset and full-start training.}
        (a) Uniform stage-reset sampling allocates equal rollout groups across stages, while informative-group counts vary.
        (b) Full-start training with stage-wise rewards produces more informative groups per update.
        (c) Shorter stage-local rollouts reduce update time, yielding $1.58\times$ the hourly informative-group throughput of full-start training.
    }
    \label{fig:signal_efficiency}
    \vspace{-4mm}
\end{figure}

\subsubsection{Effect of Stage-Training Allocation}

To isolate the effect of stage allocation, we compare the two
stage-reset variants while keeping the world model, stage-specific
rewards, local rollout horizons, GRPO configuration, and reset-state
pools unchanged.
Arrival-matched sampling follows the empirical stage-arrival
distribution of full-start rollouts, whereas uniform sampling assigns equal probability to all 4 stages.

The two variants achieve similar performance on the early and
intermediate stages, but differ substantially at the final stage:
Stage~4 completion increases from 4/18 (22.2\%) under arrival-matched
sampling to 15/16 (93.8\%) under uniform sampling.
The latter sacrifices only small amounts of Stage~2--3 completion
while producing a substantially more balanced stage-completion profile.

These results show that reproducing the naturally imbalanced visitation distribution is undesirable for stage-local policy optimization.
Uniform stage sampling explicitly reallocates imagined interaction
toward underrepresented later-stage skills, which is critical for
converting strong intermediate-stage performance into end-to-end task success.

\subsubsection{Optimization Efficiency of Stage-Local Training}
We further compare the optimization efficiency of stage-local and
full-start training with stage-wise rewards.
As shown in Fig.~\ref{fig:signal_efficiency}, uniform stage
sampling allocates equal rollout groups across the 4 stages, while
the number of informative groups naturally concentrates on stages that
remain difficult.
Although full-start training produces more informative groups per
policy update (60.8 vs.\ 22.6), its substantially longer rollouts increase
each update from 28.4 to 121.1 minutes.
Consequently, stage-local training yields 47.6 informative groups per
hour, compared with 30.1 for full-start training, corresponding to a
$1.58\times$ higher wall-clock throughput.

\subsubsection{Comparison with KIR used in WoVR}

Finally, we compare stage-local uniform training with the
Keyframe-Initialized Rollout (KIR) protocol introduced in
WoVR~\cite{jiang2026wovrworldmodelsreliable}.
KIR shortens the generated prefix by initializing rollouts near a
task-critical region, but does not explicitly organize optimization
across the multiple skill stages of the complete task.

KIR achieves cumulative completion rates of
18/20, 13/20, 10/20, and 10/20 across Stages~1--4, corresponding
to 50\% end-to-end success.
The fact that earlier-stage performance also changes under KIR is
expected because all stages share the same policy parameters, so
suffix-level optimization can transfer to behaviors outside the
directly optimized segment.
Nevertheless, these earlier stages receive no explicit stage-balanced
supervision.

In comparison, uniform stage-local training achieves
20/20, 17/20, 16/20, and 15/20, yielding 75\% end-to-end success.
The improvement is therefore not explained by prefix shortening alone.
By combining stage-specific rewards, direct stage-entry initialization,
and explicit balancing of training opportunities across stages,
stage-local training provides a more systematic optimization protocol
for long-horizon manipulation while retaining a single shared policy
for end-to-end deployment.

\section{Conclusion}

We present WoVR+, a world-model-based RL framework for improving VLA policies with limited real-robot interaction.
WoVR+ combines a geometry-grounded multi-view world model,
temporally coherent chunk-wise exploration, and stage-local imagined training to support reliable closed-loop policy optimization.
Experiments in simulation and on real robots demonstrate improved world-model fidelity and downstream policy performance, while long-horizon analyses further show the benefits of localized rewards, shorter imagined prefixes, and balanced stage-wise training.

Despite these advantages, the current framework relies on predefined task stages and data-supported stage-entry states. In addition, chunk-wise Flow-SDE introduces cross-token correlations, while the current GRPO implementation employs token-wise surrogate policy ratios. Future work may explore automatic stage discovery, adaptive reset-state construction, and joint chunk-level optimization that explicitly accounts for temporally correlated exploration.

\ifCLASSOPTIONcaptionsoff
  \newpage
\fi



%

\appendices

\section{Validation of Command-Based Proprioception}
\label{app:command_proprio}

Equation~\ref{eq:wm_proprioception} updates proprioception by temporally shifting the absolute joint-and-gripper commands, rather than learning an additional robot-state transition model. This approximation relies on the low-level controller tracking absolute targets sufficiently closely. We therefore evaluate whether its tracking error remains bounded during long-horizon autoregressive interaction and compare it with three learned transition models using alternative action representations:
\begin{itemize}
\item \textbf{Command Shift (Ours).}
The predicted proprioception is obtained directly from the temporally shifted absolute joint targets,
\begin{equation}
    \hat{\mathbf S}_{t+1:t+H}
    =
    \mathbf A^{\mathrm{joint}}_{t:t+H-1},
    \qquad H=8,
\end{equation}

\item \textbf{Joint MLP.}
A learned transition model receives the current proprioceptive state and the absolute joint-action block. MLP predicts only a residual correction to the command-shift estimate:
\begin{equation}
    \hat{\mathbf S}_{t+1:t+H}
    =
    \mathbf A^{\mathrm{joint}}_{t:t+H-1}
    +
    f_{\theta}^{\mathrm{joint}}
    \left(
        \mathbf S_t,
        \mathbf A^{\mathrm{joint}}_{t:t+H-1}
    \right).
\end{equation}

\item \textbf{Absolute-EEF MLP.}
The model instead receives the current proprioceptive state and the absolute dual-arm EEF targets obtained through forward kinematics:
\begin{equation}
    \hat{\mathbf S}_{t+1:t+H}
    =
    f_{\theta}^{\mathrm{EEF}}
    \left(
        \mathbf S_t,
        \mathbf E^{\mathrm{abs}}_{t:t+H-1}
    \right).
\end{equation}

\item \textbf{Delta-EEF MLP.}
Following the delta-EEF action interface used in WoVR~\cite{jiang2026wovrworldmodelsreliable}, the final baseline conditions the same architecture on relative EEF motion. 
\begin{equation}
    \hat{\mathbf S}_{t+1:t+H}
    =
    f_{\theta}^{\Delta\mathrm{EEF}}
    \left(
        \mathbf S_t,
        \Delta\mathbf E_{t:t+H-1}
    \right).
\end{equation}
\end{itemize}

The three learned baselines use the same MLP architecture. Using real-robot data, starting from the ground-truth state, we recursively predict eight-step blocks for 200 control steps, feeding each predicted endpoint into the next block. We evaluate 75 held-out in-distribution trajectories and 75 trajectories from two unseen tasks using NRMSE:
\begin{equation}
\mathrm{NRMSE}(k)
=
\sqrt{
\frac{1}{ND}
\sum_{i=1}^{N}
\sum_{d=1}^{D}
\left(
\frac{\hat{s}_{i,k,d}-s_{i,k,d}}{\sigma_d}
\right)^2
}.
\end{equation}

\begin{figure}[t]
\centering
\includegraphics[width=\linewidth]{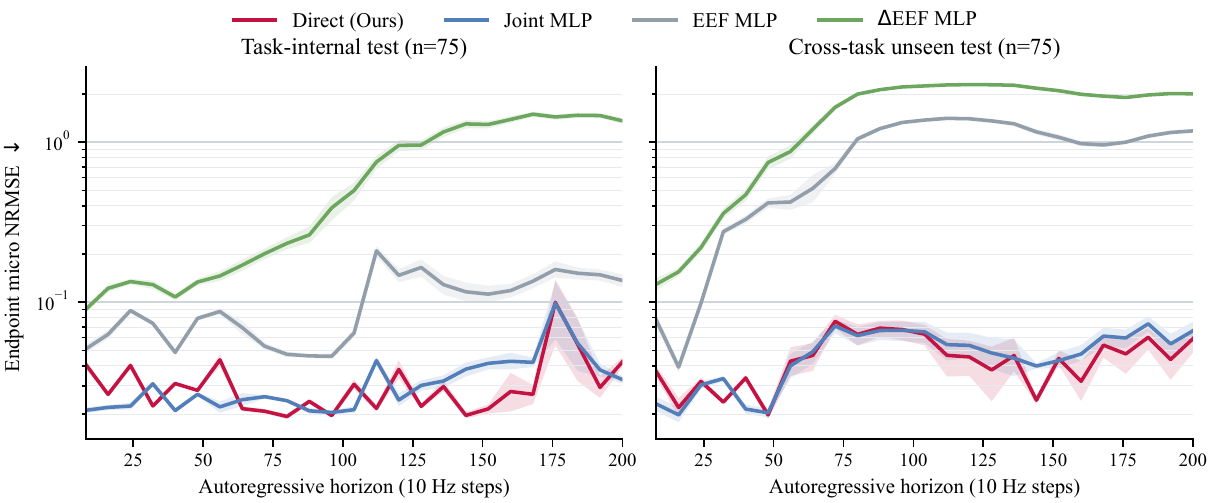}
\caption{\textbf{Long-horizon proprioception prediction.}
Endpoint NRMSE over 75 task-internal and 75 cross-task trajectories. Predictions are recursively propagated in eight-step blocks for 200 control steps.}
\label{fig:proprioception approximation}
\vspace{-3mm}
\end{figure}

Fig.~\ref{fig:proprioception approximation} reports NRMSE over all trajectories and 14 proprioceptive dimensions; shading denotes 95\% trajectory-bootstrap confidence intervals.

Joint MLP achieves the lowest task-internal error at 200 steps, showing that learned residuals can capture task-specific tracking offsets. Under cross-task shift, however, Command Shift performs best and remains substantially more stable than EEF MLP and $\Delta$EEF MLP. Thus, directly shifting absolute joint commands provides a simple and robust proprioceptive update without introducing an additional transition model whose errors can accumulate during autoregressive interaction.

\section{Reward Model used in WM-RL}
\label{app:rm}
We use lightweight task-specific ResNet classifiers to convert the current multi-view observation into a binary success reward. Synchronized head, left-wrist, and right-wrist images are resized and concatenated vertically, and processed without actions, proprioception, or observation history. We train one classifier for each task or phone-packing stage, selected by the task instruction or reset-stage identifier, respectively. Frames from the first annotated success onward are labeled positive; preceding frames and all frames from failed trajectories are labeled negative. We use inverse-class-frequency sampling and latch the reward once success is detected. All reward models remain frozen during WM-RL.

\section{Real-Robot Platform and Data Interface}
\label{app:real_robot_platform}

\begin{figure}[ht]
\centering
\includegraphics[width=\linewidth]{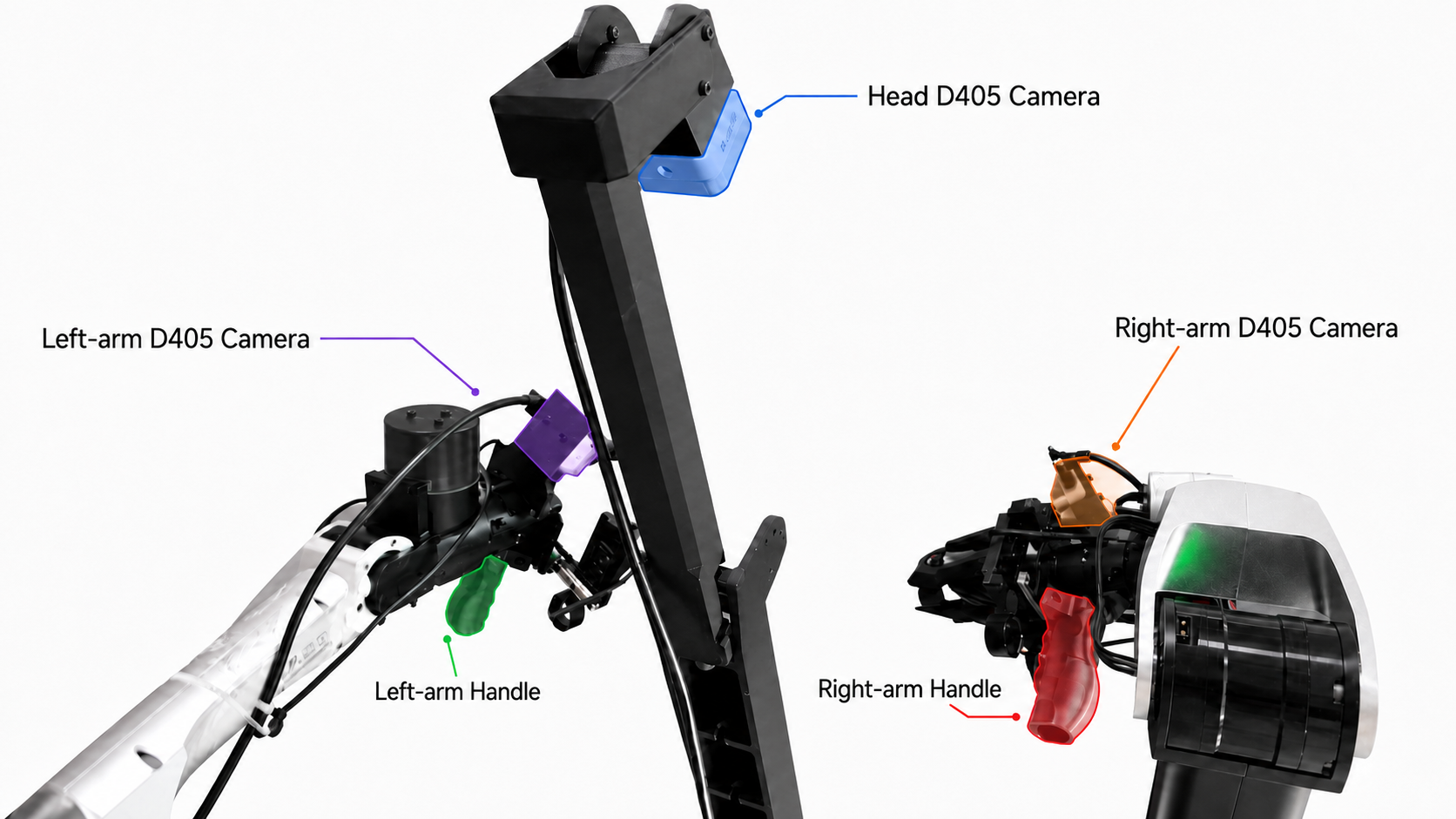}
\caption{\textbf{Dual-arm ARX-5 manipulation platform.}}
\label{fig:arxx5_platform}
\vspace{-3mm}
\end{figure}

All real-robot experiments are conducted on a dual-arm ARX-5
manipulation platform. Each arm contains six actuated joints and a parallel-jaw gripper, resulting in a 14-D joint-and-gripper proprioceptive state.
The platform is equipped with three Intel RealSense D405 cameras as shown in Fig.~\ref{fig:arxx5_platform}.

During demonstration collection, all streams retain their original timestamps and are synchronized and resampled onto the 10-Hz control timeline used for policy training and deployment.
Consequently, each policy observation consists of synchronized
head, left-wrist, and right-wrist RGB images together with the 14-D proprioceptive state.

Our demonstrations are collected through direct kinesthetic teaching. Since this procedure records the realized robot proprioception rather than an independently logged command sequence, we construct the absolute joint-and-gripper action label at time $t$ from the subsequent proprioceptive state,
\begin{equation}
    \mathbf{a}_t := \mathbf{s}_{t+1}.
    \label{eq:real_robot_action_label}
\end{equation}
For an action-chunk horizon $H$, the supervised target is therefore
\begin{equation}
    \mathbf{A}_t =
    \left(
    \mathbf{s}_{t+1},
    \mathbf{s}_{t+2},
    \ldots,
    \mathbf{s}_{t+H}
    \right).
    \label{eq:real_robot_action_chunk}
\end{equation}
This labeling convention matches the absolute-position action
interface used during autonomous policy execution and the
command-based proprioceptive update adopted by WoVR+.

\section{Task Definitions and Success Criteria}
\label{app:task_desc}
We use explicit completion criteria for all real-robot evaluations.
Table~\ref{tab:real_robot_task_definition} summarizes the six
manipulation tasks used for multi-task WM-RL, covering both
single-arm and bimanual skills.
For the long-horizon evaluation, Phone Packaging is further
decomposed into 4 sequential stages according to observable
subgoal completion. Each stage is associated with a local completion
criterion used for reward routing and stage-wise analysis, as
summarized in Table~\ref{tab:phone_packaging_stage_definition}.

\begin{table}[!t]
    \centering
    \caption{Definitions and success criteria of the six real-robot manipulation tasks.}
    \label{tab:real_robot_task_definition}

    \scriptsize
    \setlength{\tabcolsep}{3.0pt}
    \renewcommand{\arraystretch}{1.08}

    \begin{tabular}{@{}p{0.24\columnwidth}
                        p{0.34\columnwidth}
                        p{0.34\columnwidth}@{}}
        \toprule
        \textbf{Task}
        & \textbf{Objective}
        & \textbf{Success Criterion} \\
        \midrule

        Button Pressing\\(Single)
        & Press the target button.
        & The target button is successfully actuated. \\

        Bottle Righting\\(Single)
        & Grasp and reorient a fallen bottle upright.
        & The bottle remains stably upright. \\

        Test-Tube Transfer\\(Single)
        & Transfer a test tube between racks.
        & The tube is inserted into the target rack. \\

        Flower Insertion\\(Bimanual)
        & Insert two flower bundles into the vase.
        & Both bundles are placed inside the vase. \\

        Cup Stacking\\(Bimanual)
        & Arrange three cups into the target stack.
        & All cups form the designated stack. \\

        Towel Folding\\(Bimanual)
        & Fold the towel into the target shape.
        & The towel reaches the target folded configuration. \\

        \bottomrule
    \end{tabular}
\end{table}

\begin{table}[!t]
    \centering
    \caption{Stage definitions and completion criteria for the long-horizon Phone Packaging task.}
    \label{tab:phone_packaging_stage_definition}

    \scriptsize
    \setlength{\tabcolsep}{3.0pt}
    \renewcommand{\arraystretch}{1.08}

    \begin{tabular}{@{}p{0.25\columnwidth}
                        p{0.34\columnwidth}
                        p{0.33\columnwidth}@{}}
        \toprule
        \textbf{Stage}
        & \textbf{Operation}
        & \textbf{Completion Criterion} \\
        \midrule

        Stage 1\\Base positioning
        & Drag the box base to the target position.
        & The base reaches the required pose for phone placement. \\

        Stage 2\\Phone placement
        & Grasp and place the phone into the base.
        & The phone is completely inside the base. \\

        Stage 3\\Lid handover
        & Reorient the lid and transfer it to the right gripper.
        & The lid is correctly oriented and held by the right gripper. \\

        Stage 4\\Lid closure
        & Place the lid and perform bimanual fine adjustment.
        & The lid is aligned and the box is fully closed. \\

        \bottomrule
    \end{tabular}
\end{table}

\section{Training Configurations}
\label{app:training_config}

We provide the main training and implementation configurations for
both the world model and real-robot WM-RL experiments.
Table~\ref{tab:wm_training_config} summarizes the architecture and
optimization settings of the three-view world model, while
Table~\ref{tab:real_robot_wmrl_config} reports the shared policy and
WM-RL settings together with the task-specific rollout configurations.

\begin{table}[!htbp]
    \centering
    \caption{
        Architecture and training hyperparameters of the three-view world model.
    }
    \label{tab:wm_training_config}

    \scriptsize
    \setlength{\tabcolsep}{3.0pt}
    \renewcommand{\arraystretch}{1.00}

    \begin{tabular}{@{}p{0.30\columnwidth}
                        p{0.64\columnwidth}@{}}
        \toprule
        \textbf{Hyperparameter} & \textbf{Configuration} \\
        \midrule

        \multicolumn{2}{@{}l}{\textit{Architecture and conditioning}} \\

        Video backbone
        & Wan2.2-TI2V-5B \\

        Temporal window
        & 13 frames: 5 context + 8 predicted frames \\

        VACE branch
        & 15 blocks initialized from main-DiT blocks
        $\{0,2,\ldots,28\}$ \\

        Trainable modules
        & Main DiT and VACE branch; VAE and UMT5 are frozen \\

        \midrule
        \multicolumn{2}{@{}l}{\textit{Optimization and distributed training}} \\

        Scheduler
        & \texttt{FlowMatchingScheduler}, shift $5$,
        $\sigma' = 5\sigma/(1+4\sigma)$ \\

        Optimizer
        & AdamW, $\beta_1=0.9$, $\beta_2=0.999$,
        $\epsilon=10^{-8}$, LR $=10^{-5}$, weight decay $=0.01$ \\

        LR scheduler
        & \texttt{ConstantLR}, factor $1/3$ for the first
        5 optimizer steps \\

        Precision / sharding
        & bfloat16 mixed precision; DeepSpeed ZeRO-2 \\

        Batch size
        & global batch size $=16$ \\

        \bottomrule
    \end{tabular}
\end{table}

\begin{table}[!htbp]
    \centering
    \caption{
        Main configurations for the real-robot WM-RL experiments.
    }
    \label{tab:real_robot_wmrl_config}

    \scriptsize
    \setlength{\tabcolsep}{3.0pt}
    \renewcommand{\arraystretch}{1.00}

    \begin{tabular}{@{}p{0.30\columnwidth}
                        p{0.64\columnwidth}@{}}
        \toprule
        \textbf{Configuration} & \textbf{Setting} \\
        \midrule

        \multicolumn{2}{@{}l}{\textit{Shared policy and exploration}} \\

        Policy observation
        & 3 RGB views, 14-D proprioception, and instruction \\

        Policy action
        & 14-D absolute joint-and-gripper action; horizon $H=16$ \\

        Policy sampling
        & 5 flow steps; chunk-wise Flow-SDE noise at 1 sampled step \\

        Noise scale
        & $0.2$ \\

        Trainable module
        & Action expert only; no value head \\

        \midrule
        \multicolumn{2}{@{}l}{\textit{Shared WM-RL optimization}} \\

        GRPO group size
        & $8$ \\

        Imagined rollouts
        & $512$ trajectories per update \\

        Actor optimizer
        & AdamW, learning rate $5\times10^{-6}$ \\

        Compute resources
        & $8\times$ NVIDIA A100 GPUs \\
        
        \midrule
        \multicolumn{2}{@{}l}{\textit{Six-task WM-RL}} \\

        Reset pool
        & $120$ episode-initial histories per task \\

        Sampling
        & Uniform over six tasks \\

        Reward routing
        & One task-specific three-view reward model per task \\

        \midrule
        \multicolumn{2}{@{}l}{\textit{Phone Packaging}} \\

        Reset pool
        & Stage-entry histories from 100 shared reference trajectories,
        with pool size determined by stage arrival \\

        Sampling
        & Uniform over 4 stages \\

        Stage horizons
        & $144/256/240/256$ frames for Stages 1--4 \\

        Reward routing
        & One stage-specific three-view reward model per stage \\

        \bottomrule
    \end{tabular}
\end{table}

\bibliographystyle{IEEEtran}
\bibliography{references}

\end{document}